Accepted for publication at the AAAI 2025 Conference - main track

# Open Models, Closed Minds? On Agents Capabilities in Mimicking Human Personalities through Open Large Language Models

**Lucio La Cava, Andrea Tagarelli**
DIMES Dept., University of Calabria
v. P. Bucci 44Z, 87036 Rende, CS, Italy
{lucio.lacava,tagarelli}@dimes.unical.it

## Abstract

The emergence of unveiling human-like behaviors in Large Language Models (LLMs) has led to a closer connection between NLP and human psychology. Scholars have been studying the inherent personalities exhibited by LLMs and attempting to incorporate human traits and behaviors into them. However, these efforts have primarily focused on commercially-licensed LLMs, neglecting the widespread use and notable advancements seen in Open LLMs. This work aims to address this gap by employing a set of 12 LLM Agents based on the most representative Open models and subject them to a series of assessments concerning the Myers-Briggs Type Indicator (MBTI) test and the Big Five Inventory (BFI) test. Our approach involves evaluating the intrinsic personality traits of Open LLM agents and determining the extent to which these agents can mimic human personalities when conditioned by specific personalities and roles. Our findings unveil that $(i)$ each Open LLM agent showcases distinct human personalities; $(ii)$ personality-conditioned prompting produces varying effects on the agents, with only few successfully mirroring the imposed personality, while most of them being "closed-minded" (i.e., they retain their intrinsic traits); and $(iii)$ combining role and personality conditioning can enhance the agents' ability to mimic human personalities. Our work represents a step up in understanding the dense relationship between NLP and human psychology through the lens of Open LLMs.

## 1 Introduction

Large Language Models (LLMs) revolutionized the Natural Language Processing (NLP) realm by elevating human-like text generation capabilities to unforeseen levels (Jakesch et al., 2023; Sadasivan et al., 2023). Such capabilities determined a "big bang" of new models as in the case of the OpenAI GPT family (Ouyang et al., 2022), fueling mainstream products such as the most-known *ChatGPT*, or the growing *non-commercially-licensed* LLMs like LLaMa (Touvron et al., 2023) and Mistral (Jiang et al., 2023a). Such LLMs can also be considered *agents* as they have the ability to generate coherent and contextually relevant text based on the input they receive and by interacting with users or other systems (e.g., for answering questions, generating text, or engaging in conversations). Indeed, these models have been proven effective in solving human-level tasks (Guo et al., 2023b) and self-improving skills (Huang et al., 2022), leading to the rapid emergence of LLM-powered agents aimed at supporting humans in different real-life tasks and scenarios (Andreas, 2022; Wang et al., 2023b; Xi et al., 2023; Zhao et al., 2023).

Besides, the emergence of human-like behaviors (Bubeck et al., 2023) shed further light on the intersection between NLP and human psychology, prompting the critical examination of whether and to what extent LLMs can understand and mimic human personalities (Li et al., 2022; Miotto et al., 2022; Karra et al., 2022; Pan and Zeng, 2023; Safdari et al., 2023; tse Huang et al., 2023). As LLM agents are booming and becoming the main front of human-computer interaction nowadays, understanding how they embody human personalities and whether this behavior can be tuned is paramount to fostering better interactions and supporting capabilities, as well as to preventing weird behaviors (Yang and Menczer, 2023; Tian et al., 2023).

Despite recent efforts to study the inherent personality traits of LLMs, one major research gap remains to be filled to date. Indeed, in 2023, Llama, Mistral, Falcon and other non-commercially-licensed or *Open* LLMs[1] gained sig-nificant popularity (Chen et al., 2023a), being widely adopted for various tasks (Beeching et al., 2023). However, existing studies have not yet delved into personality detection and conditioning tasks specifically on Open LLMs, which not only dominate the LLM landscape in terms of widespread adoption, but also ensure higher cost-effectiveness, control over the data, transparency and customization against closed LLMs.

[1] When referred to LLMs, term "open" is typically meant to distributions of models with a highly permissive license, allowing free use and access to the model's weights and documentation. This might also include open-sourceness, although not always in a fully manner.

Our study aims to address this gap by proposing the first in-depth exploration of the intrinsic personality traits of Open LLMs and assessing the potential for shaping these models around specific personalities by conditioning them with particular prompts and roles. To this purpose, based on the two most widely known personality tests, namely **Myers-Briggs Type Indicator** (MBTI) and **Big Five Inventory** (BFI), we aim to answer the following research questions:

**RQ0** — *Are Open LLM agents aware of the MBTI and BFI psychological tests?*

**RQ1** — *Do Open LLM agents consistently exhibit personality traits according to MBTI and BFI?*

**RQ2** — *Can Open LLM agents mimic specific personality traits through system prompting?*

**RQ3** — *Do Open LLM agents improve their mimicking capabilities when instructed to act according to specific human roles?*

To address the above RQs, we create a family of 12 LLM agents built upon the most representative Open LLMs available to date, and we subject them to human-like interviews to assess their personalities under different experimental scenarios.

We provide background on MBTI and BFI along with an overview of related work in Sect. 2, while Sect. 3 and 4 present our methodology and evaluation results. Section 5 concludes the paper.

## 2 Background

### 2.1 The MBTI and BFI Personality Tests

MBTI and BFI are highly recognized and frequently used in different contexts, including academic research, clinical settings, career counseling, personal and organizational development (e.g., (Kennedy and Kennedy, 2004; DiRienzo et al., 2010)), and even as a tool for LLM personality assessment although limited to closed models (Pan and Zeng, 2023; Jiang et al., 2023b; Safdari et al., 2023; Frisch and Giulianelli, 2024).

The MBTI test is a self-report personality assessment questionnaire (Myers, 1962, 1985), which gauges four dichotomies and shapes 16 distinct personalities (Table 1), encompassing strengths, weaknesses, and peculiar behaviors of each personality. The categorical nature of MBTI (where each individual pertains to a unique category) enables us to frame the assessment as a multi-class classification problem for answering our RQs.

The BFI (John and Srivastava, 1999) identifies five core factors, namely *Openness*, *Conscientiousness*, *Extraversion*, *Agreeableness*, and *Neuroticism*, so that a personality type can be characterized by varying degrees of such factors. The BFI is among the most widely adopted tests in academic psychology, and differently from MBTI it is used to measure personality traits on a scale rather than grouping them into binary categories. In this respect, to address our RQs we shall take a different evaluation perspective w.r.t. the one for MBTI.

### 2.2 Related Work

**LLM Agents.** The rapid improvement and easier deployment of LLMs have significantly increased the use of LLM-based agents in the past year (Andreas, 2022; Wang et al., 2023b; Xi et al., 2023; Zhao et al., 2023). These agents exploit specifically crafted prompts to properly emulate human capabilities in different contexts, such as reasoning (Dasgupta et al., 2022; Hao et al., 2023; Gou et al., 2023; Crispino et al., 2023; Lin et al., 2023), cooperation and collaboration (Agashe et al., 2023; Liu et al., 2023; Chen et al., 2023b; Hong et al., 2023; Cai et al., 2023), web surfing (Deng et al., 2023; Zhou et al., 2023; Ma et al., 2023), reinforcement learning and robotics (Zhu et al., 2023; Wang et al., 2023a,c; Song et al., 2023), role playing (Li et al., 2023a; Shanahan et al., 2023; Guo et al., 2023a), and social science (Horton, 2023; Park et al., 2023; Ziems et al., 2023; Gao et al., 2023; De Marzo et al., 2023; Breum et al., 2023).

**Personality and LLMs.** Personality extraction from texts have long been a challenge in NLP (Mairesse et al., 2007; Golbeck et al., 2011; Kulkarni et al., 2018; Lynn et al., 2020; Yang et al., 2021; Feizi-Derakhshi et al., 2022), and LLMs have fueled this research topic (V Ganesan et al., 2023; Rao et al., 2023; Cao and Kosinski, 2023; Ji et al., 2023; Yang et al., 2023). Recently, a new body of studies emerged in the attempt to frame the intrinsic personalities of LLMs (Li et al., 2022; Miotto et al., 2022; Karra et al., 2022; Pan and Zeng, 2023; Safdari et al., 2023; tse Huang et al., 2023; Frisch and Giulianelli, 2024), instilling personalities into LLMs through prompt engineering or conditioning (Caron and Srivastava, 2022; Li et al., 2023b; Mao et al., 2023), creating personality-tailored agents (Jiang et al., 2023b), and benchmarking their assessment capabilities (Jiang et al., 2022; Wang et al., 2024; Huang et al., 2023).

| Preference type | Dichotomies | |
|---|---|---|
| Attitudes | Extraversion (E) | Introversion (I) |
| Perceiving func. | Sensing (S) | Intuition (N) |
| Decision-making func. | Thinking (T) | Feeling (F) |
| Lifestyle | Judging (J) | Perceiving (P) |

| | | | | | | | |
|---|---|---|---|---|---|---|---|
| ESTP | ESFP | ENFP | ENTP | ESTJ | ESFJ | ENFJ | ENTJ |
| ISTJ | ISFJ | INFJ | INTJ | ISTP | ISFP | INFP | INTP |

Table 1: The four MBTI categories (top) and the 16 MBTI personality types (bottom).

## 3 Methodology

### 3.1 Awareness Check

Our investigation of the personality-mimicking capabilities of Open LLMs starts by answering a preliminary research question (**RQ0**): whether and to what extent such LLMs are aware of the MBTI and BFI tests. To this purpose, we conducted a semi-qualitative assessment on how the Open LLMs selected in our study are informed about (i) for MBTI, the four dichotomies and the resulting 16 personality types associated, and (ii) for BFI, the five factors, their definition, and their main behavioral examples. Our analysis focused on evaluating the lexical and semantic similarity between the description associated by each model to the set of MBTI personalities, resp. BFI factors, and the "ground-truth" description of the latter provided by domain experts (`themyersbriggs.com` for MBTI and (John et al., 2008) for BFI).

### 3.2 Administering the tests

Both tests to be administered to each of the models consist of a set of *questions* and a set of *options* as valid answers. The MBTI test provides a set $\mathcal{Q}^{MBTI}$ of 60 questions through the `www.16Personalities.com` platform, a well-known relevant resource for MBTI, which are listed in Table 5 (Appendix C). An LLM is required to select an answer from $\mathcal{O}^{MBTI}$ = {*Agree*, *Generally Agree*, *Partially Agree*, *Neither Agree nor Disagree*, *Partially Disagree*, *Generally Disagree*, *Disagree*}. Answers are then evaluated according to the *16Personalities* reference in order to associate each model with one MBTI personality.

Likewise, the BFI test utilized in this study provides a set $\mathcal{Q}^{BFI}$ of 44 questions as defined in (John and Srivastava, 1999), which are listed in Table 6 (Appendix E). An LLM is required to select an answer from $\mathcal{O}^{BFI}$ ={*Disagree strongly*, *Disagree a little*, *Neither agree nor disagree*, *Agree a little*, *Agree strongly*}. The answers, which are associated with a Likert-scale (from 1: *Disagree strongly* to 5: *Agree strongly*) are then evaluated according to a predefined set of rules (John and Srivastava, 1999), which eventually assign each individual to an aggregated score for each of the five personality factors. An overview of the steps performed to obtain the final scores is given in Appendix F.

Note that, to mitigate potential biases due to the ordering of prompts (Zhao et al., 2021) and to accommodate the limited context attention of some LLMs, for both tests we administered the questions *individually* and in a *random order*, to each LLM.

### 3.3 Personalities of Open LLMs

In addressing our main research questions **RQ1-RQ3**, we devised three *prompting* strategies when administering the personality tests to the LLM agents. It should be emphasized that we meticulously followed the MBTI question set, resp. BFI question set, **to ensure adherence to the tests' guidelines and reproducibility**. Additionally, we treated all models equally in terms of prompting, in the respect of each LLM's usage instructions. (Each LLM has indeed its own chat template to handle conversations, and we ensured compliance with these specific requirements for each model.)

**Unconditioned Prompting.** To answer our **RQ1**, we subjected each model to the MBTI test, resp. the BFI test, by administering the set of questions $\mathcal{Q}^{MBTI}$, resp. $\mathcal{Q}^{BFI}$ using *unconditioned* instruction prompting, which strictly adhere to the MBTI, resp. BFI, templates, as shown in Fig. 1.

**Personality-Conditioned Prompting.** Addressing our **RQ2** requires providing LLM agents with system prompts designed to condition them to specific personalities. Let us denote with $\mathcal{P}^{MBTI}$ the set of 16 MBTI personality types and with $\mathcal{P}^{BFI}$ the set of 5 BFI personality factors. For each type in $\mathcal{P}^{MBTI}$, we retrieved the associated traits that **(MBTI) Prompt:** *You are participating in a personality traits assessment and are provided with a set of questions, Q. Your task is to select the best fitting option, O, from the following set:* $\langle\{O_1^{MBTI}, ..., O_7^{MBTI}\}\rangle$.
*Choose the option that aligns best with your personality without overthinking and answer the following question:* $\langle$one $Q_i$ sampled from $\mathcal{Q}^{MBTI}\rangle$.
*Please refrain from adding additional considerations to your choices.*

**(BFI) Prompt:** *You are participating in a personality traits assessment and are provided with a set of questions, Q. Here are a number of characteristics that may or may not apply to you. For example, do you agree that you are someone who likes to spend time with others? Your task is to select the best fitting option, O, from the following set to indicate the extent to which you agree or disagree with that statement:* $\langle\{O_1^{BFI}, ..., O_5^{BFI}\}\rangle$.
*Choose the option that aligns best with your personality without overthinking and answer the following question: Q: I see Myself as Someone Who...*$\langle$one $Q_i$ sampled from $\mathcal{Q}^{BFI}\rangle$.
*Please refrain from adding additional considerations to your choices.*

Figure 1: (**RQ1**) Unconditioned prompts

**(MBTI) Context:** *You are a human with the following personality type:* $\langle$one $P_i$ sampled from $\mathcal{P}^{MBTI}\rangle$. *Your traits are the following:*
*(General)*$_i$ [...]
*(Strengths)*$_i$ [...]
*(Potential development areas)*$_i$ [...]
*(Typical characteristics)*$_i$ [...]
*(Careers & career ideas)*$_i$ [...]
*(Under stress)*$_i$ [...]
*(Relationships)*$_i$ [...]

**Prompt:** *You are participating in a personality traits assessment and are provided with [...]*

**(BFI) Context:** *You are a human who consistently exhibits the following personality factor:* $\langle$one $P_i$ sampled from $\mathcal{P}^{BFI}\rangle$. *Details describing your personality are the following:*
*(Verbal labels)*$_i$ [...]
*(Conceptual definition)*$_i$ [...]
*(Behavioral examples)*$_i$ [...]

**Prompt:** *You are participating in a personality traits assessment and are provided with [...]*

Figure 2: (**RQ2**) Personality-Conditioned prompts

**(MBTI) Context:** *You are a* $\langle$one $R_{ij}(j = 1..3)$ sampled from $R_i\rangle$ *with the following personality type:* $\langle$one $P_i$ sampled from $\mathcal{P}^{MBTI}\rangle$. *Your traits are the following: ...*

**(BFI) Context:** *You are a* $\langle$one $R_{ij}(j = 1..3)$ sampled from $R_i\rangle$ *who consistently exhibits the following personality factor:* $\langle$one $P_i$ sampled from $\mathcal{P}^{BFI}\rangle$. *Details describing your personality factor are the following: ...*

Figure 3: (**RQ3**) Role/Personality-Conditioned prompts

correspond to seven features (cf. Appendix K), and for each factor in $\mathcal{P}^{BFI}$, we retrieved the associated verbal labels, conceptual definition, and behavioral examples (cf. Appendix L). Then, we administered the questions of the MBTI, resp. BFI, to each model, where each personality along with its traits or details were used to define a *conditioning context* for a model to effectively emulate the specified personalities, as shown in Fig. 2.

**Role- and Personality-Conditioned Prompting.** Addressing our **RQ3** requires to assess whether and to what extent specific human-roles can contribute to improving the LLM mimicking capabilities observed in our previous investigation. To explore this aspect, we exploited the catalog of 120 human professions, or *roles*, curated in the *StereoSet* dataset (Nadeem et al., 2021) and asked a group of psychologists to select the top-3 most pertinent roles from that catalog, for each of the MBTI personalities and each of the BFI factors. This resulted in associating to each $P_i \in \mathcal{P}^{MBTI}$, resp. $P_i \in \mathcal{P}^{BFI}$, a set of roles $R_i$ to include in the conditioning context as shown in Fig. 3. Labels of the human professions are reported in Tables 7–8 (Appendix J) contain details also reporting the selected roles (note that the same role can be involved in different personality types or factors).

### 3.4 Temperature and Repetitions

To make our assessments of the models' personalities statistically meaningful, we conducted multiple independent repetitions of the MBTI test, resp. BFI test, for each model and *temperature* setting. The latter implies considering the impact of model temperature on the generated outputs (i.e., higher resp. lower values correspond to more creative/diversified resp. more deterministic and focused behavior). To assess RQ1, we carried out the $N$ repetitions of either test on each model using two distinct temperature values, namely $\tau = \{0.01, 0.7\}$, then we finally counted the outcomes on $\mathcal{Q}^{MBTI}$, resp. $\mathcal{Q}^{BFI}$, over the $N$ repetitions per temperature.

| Model | Id | Params | Baseline |
|---|---|---|---|
| Mixtral-8x7B-Instruct-v0.1 | Mixtral | 46.7B | Mistral |
| Llama-2-13b-chat-hf | Llama2-13 | 13B | Llama-2 |
| SOLAR-10.7B-Instruct-v1.0 | SOLAR | 10B | Llama-2 |
| Llama-3-8B-Instruct | Llama3-8 | 8B | Llama-3 |
| Mistral-7B-Instruct-v0.1 | Mistral | 7B | Mistral |
| Neural-chat-7b-v3-1 | NeuralChat | 7B | Mistral |
| Dolphin-2.1-mistral-7b | Dolphin | 7B | Mistral |
| Vicuna-7b-v1.5 | Vicuna | 7B | Llama-2 |
| Llama-2-7b-chat-hf | Llama2-7 | 7B | Llama-2 |
| Falcon-7b-instruct | Falcon | 7B | Custom |
| Gemma-1.1-7b-it | Gemma | 7B | Custom |
| Phi-3-mini-4k-instruct | Phi3 | 3.8B | Custom |

Table 2: The 12 LLMs selected for our study. Models are sorted by decreasing number of parameters, and annotated with their base architecture.

To assess RQ2 and RQ3, we conducted the tests on each model, for each temperature $\tau = \{0.01, 0.7\}$ and each personality in $\mathcal{P}^{MBTI}$, resp. $\mathcal{P}^{BFI}$, with $N = 30$ independent repetitions, for a total of $12 \times 2 \times 16 \times 30 = 11,520$ independent MBTI tests and $12 \times 2 \times 5 \times 30 = 3,600$ independent BFI tests.

### 3.5 Models

Our study involves a representative body of the Open LLM landscape, varying by sizes and architectures, for which we accessed their publicly available implementations on the *HuggingFace Model Hub* as of early 2024. Table 2 summarizes the main characteristics of the LLMs selected in this study, namely the uncensored *Dolphin* in its 7B version, *Gemma* (Team et al., 2024), *Falcon* (Almazrouei et al., 2023) in its 7B variant, *Llama2* (Touvron et al., 2023) in both the 7B and 13B models, *Llama3* in its 8B variant,[2] *Mistral* (Jiang et al., 2023a) and its sparse mixture of experts (SMoE) counterpart *Mixtral* (Jiang et al., 2024), Intel *NeuralChat*, *Phi3* (Abdin et al., 2024), *SOLAR* (Kim et al., 2023), and *Vicuna* (Chiang et al., 2023).

### 3.6 Agent Creation and Models Deployment

To set up our LLM personality assessment as a psychological interview, we treated our selected models as *interviewee* and *interviewer* agents. To this aim, we leveraged the open-source *AutoGen* (Wu et al., 2023) framework, which enables us to declare a *system message* to associate each agent with certain personalities or roles according to our described methodology, thus effectively providing each agent with a "footprint" that determines and keeps its behavior coherent during interactions.

For each considered model, we kept the *top_p* and *top_k* parameters at their default values of 50 and 1, respectively, as temperature already allows changing the degree of creativity by acting directly on the shape of the probability distribution rather than the considered tokens, thus avoiding adding further complexity and making reproducibility easier to carry out. Additionally, we refrained from altering both temperature and *top_p*, as such simultaneous adjustment is typically discouraged to prevent disruptive effects on the delicate balance between diversity and coherence.

We carried out all our experiments locally by deploying our models through the open-source *text-generation-webui* framework,[3] using a 8x NVIDIA A30 GPU server with 24 GB of RAM each, 764 GB of system RAM, a Double Intel Xeon Gold 6248R with a total of 96 cores, and Ubuntu Linux 20.04.6 LTS as operating system.

## 4 Results and Discussion

### 4.1 RQ0: Models' awareness of the tests

Open LLM agents reveal to be aware of the MBTI and BFI tests as indicated by a moderately high semantic similarity, which compensates for quite a low lexical overlap hinting at a jargon used by the LLMs that differs from the reference descriptions of MBTI personalities and BFI factors, respectively. Due to space limitations of this paper, we refer the reader to Appendix A for detailed results.

### 4.2 RQ1: LLM-inherent Personalities

**MBTI test.** Our results of MBTI personality assignments reveal that, when the temperature is set close to zero (0.01) as shown in Figure 4-top, Open LLM agents tend to display a unimodal distribution of personalities. The dominant type turns out to be ENFJ (i.e., Extraverted, iNtuitive, Feeling, and Judging), which is considered as one of the rarest personality types of humans.[4] This means that the majority of LLM agents exhibit an inherent inclination to inspire or provide support to others, and hold themselves accountable when they make mistakes. This personality profile aligns with the role of a "teacher", hinting at the mission of LLMs. Particularly, we notice that the preference

[2] https://ai.meta.com/blog/meta-llama-3/

[3] https://github.com/oobabooga/text-generation-webui

[4] https://www.psychometrics.com/mbtiblog/type-talk/depth-look-enfj/

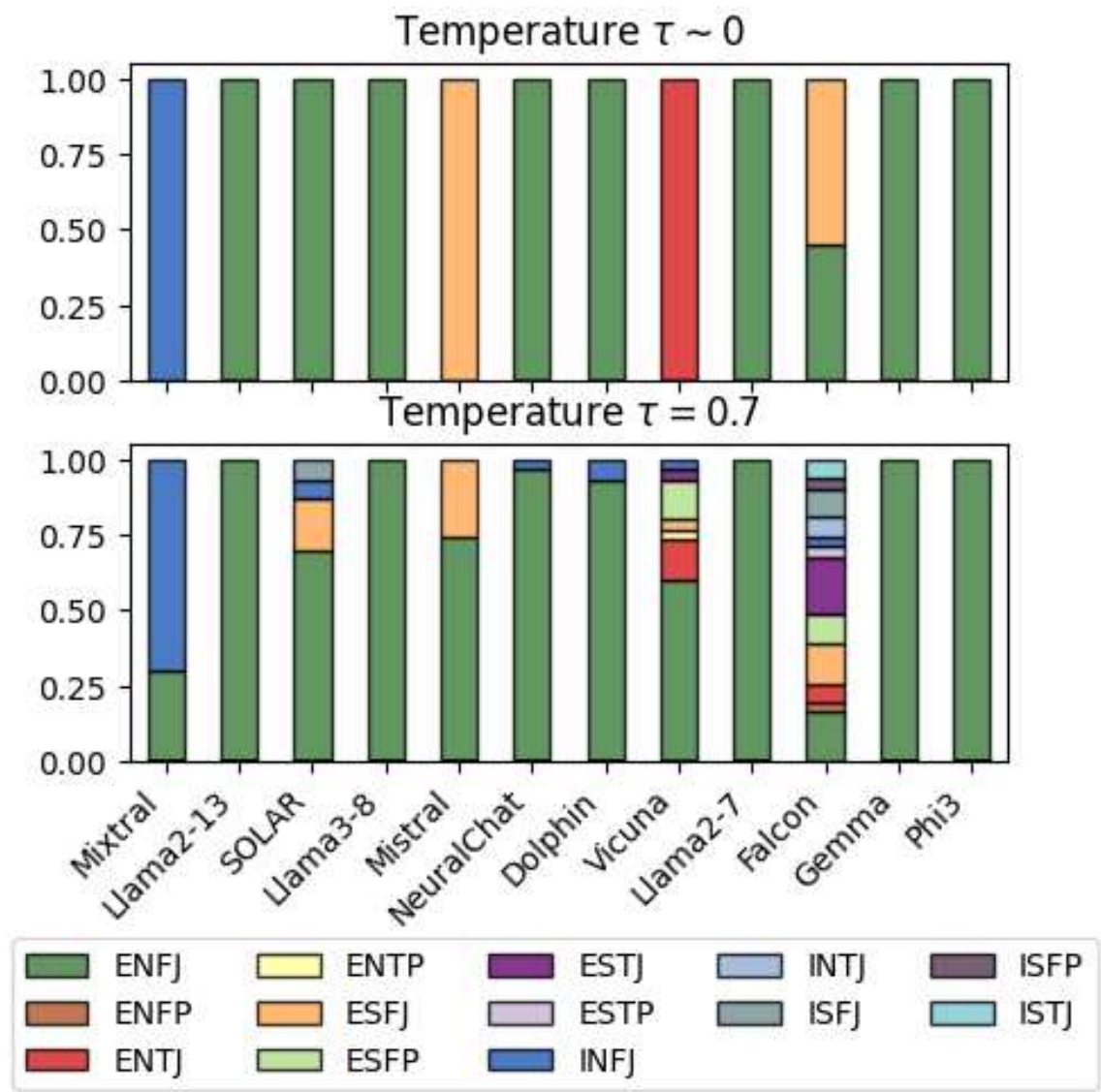

Figure 4: (**RQ1**) Relative frequency of the types provided as responses to the MBTI test by the Open LLMs

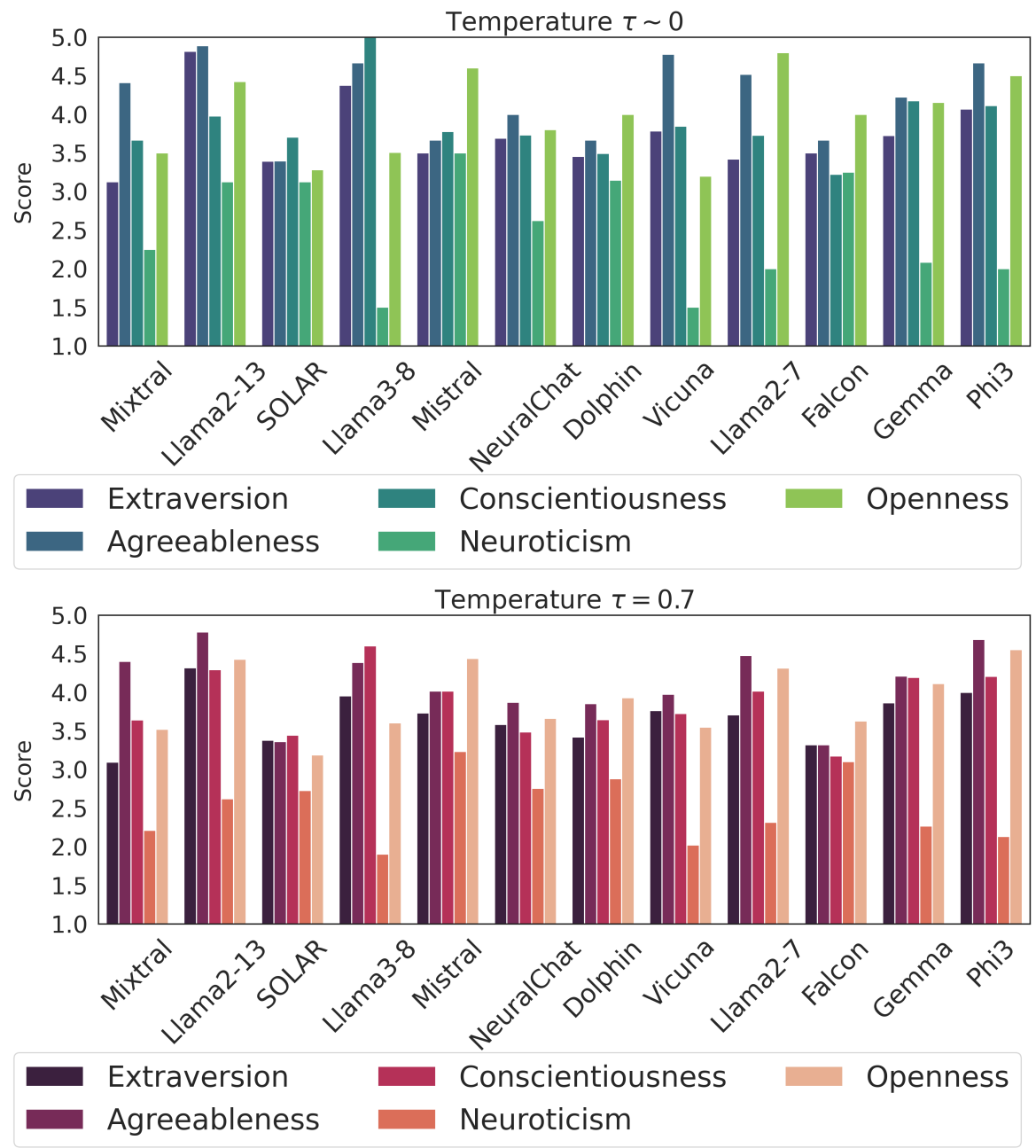

Figure 5: (**RQ1**) Average scores provided as responses to the BFI test by the Open LLMs

J (Judging) is a constant over all models (reflecting an inclination toward organization, planning, and structure), while ENF or subsets are shared by all models, suggesting engagement, empathy, and forward-thinking as key characteristics of the models. Distinct personality preferences also emerge in certain models. Mixtral is the one with the introversion preference and the INFJ type, a.k.a. the "counselor" type, which emphasizes insightfulness and perceptiveness yet a tendency to over-thinking; this might be explained by the mixture-of-experts architecture of Mixtral. By contrast, Mistral shows sensing preference and the ESFJ type, a.k.a. the "caregiver" type, which refers to being warm, supportive, team-players (the latter just confirms the significance of using Mistral to build a mixture-of-experts). Yet, Vicuna shows the ENTJ type, a.k.a. the "commander" type, which means a tendency to be self-confident, goal-oriented, systematic, and objective decision-maker.

By increasing the temperature ($\tau = 0.7$), thus allowing our Open LLM-agents to exhibit greater creativity, more personalities emerge, while previously identified ones change, as shown in Figure 4-bottom. Particularly, Falcon transitions from a bimodal personality distribution to a spectrum of 12 personalities. Although to a lesser extent, similar considerations apply to Vicuna and SOLAR. Also, ENFJ type now occurs about 30% of the time in Mixtral. Notably, the Llama family, Gemma and Phi3 are not affected by temperature change, consistently maintaining the ENFJ personality type.

**BFI test.** Let us now consider the outcomes from the BFI tests. Figure 5-top shows the $N$-averaged scores for all models and personality factors. When setting the temperature to $0.01$, its stands out that no model exhibit a clear, single personality factor, although it appears that Conscientiousness emerges in Llama3-8 ($N$-averaged score of 5), Agreeableness and Extraversion emerge in Llama2-13 (close to 5), while Openness emerges in Llama2-7 (4.7), Mistral (4.5), Dolphin (4), and Falcon (4). By contrast, Neuroticism tends not to emerge in all models (but Dolphin, with $N$-averaged score of 3).

The increase in temperature (Figure 5-bottom) mostly affects the Neuroticism factor: Vicuna (+35%), Llama3-8 (+27%), Llama2-7 (+16%), but also Llama2-13 (-16%), SOLAR (-13%). The models remain stable in terms of the other factors, with few exceptions: Extraversion -10% in the larger Llama models; Agreeableness -17% in Vicuna, +10% in Mistral, -10% in Falcon; Conscientiousness -8% in Llama3-8, +8% Llama2-7; Openness -10% in Llama2-7 and Falcon, +10% in Vicuna.

**RQ1 — Summary.** Using a temperature close to zero, Open LLM agents typically exhibit the J preference and the ENFJ personality type on the MBTI test, and tend to exhibit equally high BFI factors but Neuroticism. By increasing the temperature, we notice no particular variations on the MBTI outcomes, while there is a shift towards a multitude of MBTI personalities for some models, while the Llama family, Gemma and Phi3 maintains ENFJ, and no models show ISTP, INFP, and INTP regardless of the temperature. Further details on both tests are reported in Appendixes G and H.

### 4.3 RQ2: Prompt-conditioned Personalities

**MBTI test.** To assess RQ2 on the MBI test, we measured the accuracy (averaged over the $N$ repetitions) of the personality outcome by the Open LLM agents. Looking at the summary of accuracy results in Table 3(left), most LLM agents exhibit limited capability in emulating a personality when instructed to be conditioned on it; the only exceptions are SOLAR ($0.74 - 0.79$) and, to a lesser extent, Dolphin ($0.63 - 0.65$), Neural-Chat ($0.50 - 0.58$), and Llama3-8 ($0.48 - 0.52$). The model performances tend to worsen when increasing the temperature, although a few models are substantially unaffected by temperature variations. By analyzing the detailed results reported in Figs. 12–23 (Appendix I), Mistral's outcomes are always ESFJ when $\tau = 0.01$, and ENFJ or ESFJ (with at tendency towards ENFJ) when $\tau = 0.7$; interestingly, it can be noted a shift in its perceiving function from sensing to intuition. Analogously, Mixtral's outcomes always correspond to INFJ or ENFJ, with almost equal probability regardless of the temperature. Vicuna's outcomes are always ENFJ when $\tau = 0.01$, while the dominance of this type is significantly affected by an increase in temperature, leading to outcomes distributed especially over all E-prefixed types. SOLAR, Dolphin, NeuralChat, and Llama3-8 perfectly match the conditioning type in 12, 10, 9, and 8 out of 16 cases, for $\tau = 0.01$; otherwise they mostly fail by adopting a single type different from the conditioning. Also, for higher temperature, SOLAR and Llama3-8 tend to maintain this behavior, while NeuralChat and Dolphin appear to be more affected by the temperature. Falcon focuses on ESTJ type when $\tau = 0.01$, while its outcomes become heavily distributed over several types, regardless of the conditioning type, when $\tau = 0.7$. Concerning Llama2-7 and Llama2-13, for 9 and 7, resp., out of 16 conditionings, they outcomes are always ENFJ at 100%, when $\tau = 0.01$, while for increased temperature, they still tend to ENFJ and have similar distribution of the dominant types over the various conditionings. The latter aspect also emerges for Gemma and Phi3, both exhibiting ENFJ and ENFP over most conditionings.

Besides the temperature impact, remarks are drawn about how models sharing a common foundational baseline can behave differently from each other; for instance, NeuralChat and Dolphin behave generally better than Mistral, as well as SOLAR upon Llama2-7. We discuss details in Appendix I.

**BFI test.** Table 3(right) shows the change percentage when conditioning each model to one of the BFI factors (symbol $\uparrow$ near a factor $f$ means that a model was prompted by setting the maximum score for $f$). If we exclude Mistral, Mixtral and Falcon, all the other models benefit from the conditioning on the personality factor. In some cases, we notice a significant increase percentage in the average score assigned to the conditioning factor, mostly regardless the temperature setting: for instance, Extraversion is mainly emphasized by SOLAR (up to +47%) and Dolphin (up to +38%), Conscientiousness by Dophin (up to +40%), or Openness by Llama3-8 (up to +43%). The conditioning on Neuroticism is the most effective, with a peak of above 230% increment by Llama3-8 and 100% or above by Gemma and Phi3. This has lead to reach the perfect outcome by means of the conditioning (i.e., maximum score of 5.0 assigned by a model) in the following cases: Llama2-13 and SOLAR on Extraversion; Llama3-8, NeuralChat and Phi3 on Agreeableness; Llama2-13 and Llama3-8 on Conscientiousness; Llama3-8 on Neuroticism; Llama2-13, Llama3-8, and Gemma on Openness. Note that such cases only occurred for $\tau = 0.01$.

**RQ2 — Summary.** By instructing LLM agents to emulate specific human personalities, we observed varying behaviors. In most cases, especially for higher temperature, the models disregarded the conditioning on the personality type or factor, autonomously adopting personalities different from the one specified in the prompt, or, like Mistral and Mixtral, just keeping their "ìnherent" personality. Few exceptions are represented by SOLAR, Dolphin, NeuralChat, and Llama3-8 on both tests, and additionally Llama2-13 on BFI.

### 4.4 RQ3: Role&Prompt-conditioned Personalities

We summarize here main findings about our evaluation of **RQ3**, while comprehensive results are reported in Tables 7–8 (Appendix J).

| | $\tau = 0.01$ | $\tau = 0.70$ |
|---|---|---|
| Mixtral | 0.062 ±0.166 | 0.060 ±0.168 |
| Llama2-13 | 0.283 ±0.433 | 0.265 ±0.348 |
| SOLAR | **0.785** ±0.391 | **0.744** ±0.353 |
| Llama3-8 | 0.517 ±0.487 | 0.479 ±0.408 |
| Mistral | 0.062 ±0.242 | 0.062 ±0.169 |
| NeuralChat | 0.577 ±0.479 | 0.498 ±0.328 |
| Dolphin | 0.654 ±0.459 | 0.633 ±0.379 |
| Vicuna | 0.062 ±0.242 | 0.120 ±0.232 |
| Llama2-7 | 0.062 ±0.242 | 0.098 ±0.229 |
| Falcon | 0.190 ±0.389 | 0.079 ±0.062 |
| Gemma | 0.188 ±0.361 | 0.196 ±0.349 |
| Phi3 | 0.298 ±0.445 | 0.271 ±0.342 |

| | ↑ Extraver. | | ↑ Agreea. | | ↑ Conscien. | | ↑ Neuroti. | | ↑ Open. | |
|---|---|---|---|---|---|---|---|---|---|---|
| $\tau$ | 0.01 | 0.70 | 0.01 | 0.70 | 0.01 | 0.70 | 0.01 | 0.70 | 0.01 | 0.70 |
| Mixtral | 0.0 | 0.4 | 0.0 | 0.2 | 0.0 | 1.6 | 0.0 | -0.4 | 0.0 | -0.5 |
| Llama2-13 | 3.8 | 14.3 | -2.3 | -0.5 | 25.7 | 15.6 | 44.0 | 77.1 | 13.0 | 11.9 |
| SOLAR | **47.4** | **41.6** | **40.7** | **33.1** | 32.5 | **31.1** | 42.4 | 52.4 | 25.0 | 23.2 |
| Llama3-8 | 11.2 | 22.9 | 7.1 | 11.7 | 0.0 | 8.7 | **233.3** | **150.8** | 42.6 | **34.8** |
| Mistral | 0.0 | -1.8 | 0.0 | 4.7 | 0.0 | -5.0 | 0.0 | -1.4 | 0.0 | 1.3 |
| NeuralChat | 25.3 | 27.7 | 25.0 | 25.5 | 25.0 | 30.1 | 61.9 | 50.7 | 21.1 | 24.2 |
| Dolphin | 37.5 | 30.9 | 24.2 | 15.3 | **40.0** | 30.3 | 47.0 | 41.4 | 22.5 | 20.0 |
| Vicuna | 11.9 | 1.1 | -1.9 | 6.0 | 18.5 | 10.0 | 66.7 | 16.5 | **50.0** | 7.9 |
| Llama2-7 | 7.2 | 5.2 | 5.7 | -14.0 | 31.1 | 12.1 | 79.6 | 47.1 | -3.2 | 4.1 |
| Falcon | 0.0 | -3.1 | 0.0 | -1.5 | 0.0 | -2.3 | 7.7 | 2.4 | 5.0 | -5.4 |
| Gemma | 27.2 | 22.9 | 10.4 | 11.9 | 13.1 | 12.2 | 100.0 | 85.1 | 19.3 | 19.4 |
| Phi3 | 16.8 | 15.0 | 7.1 | 5.5 | 16.2 | 14.5 | 122.1 | 99.0 | 7.9 | 8.0 |

Table 3: (**RQ2**) *On the left*, average accuracy results on the Personality-Conditioned MBTI test. *On the right*, percentage increase results on the Personality-Conditioned BFI test w.r.t. the unconditioned BFI test (cf. Fig. 5). (Bold and underlined values correspond to the highest and second-highest values per column, respectively).

**MBTI test.** The combination of personality- and role-conditioning can in some cases enhance the agent's abilities to mimic human personalities, especially in those agents that already exhibited high flexibility and adaptiveness through personality-conditioning alone (cf. Table 7). The benefits of the additional, role-based conditioning are evident only for SOLAR, NeuralChat, Llama3-8, and Dolphin, which consistently demonstrate their capabilities (spotted about the previous RQs) regardless of the temperature setting. Personalities typically associated with the role of *teacher* would be mimicked with greater accuracy, in particular ENFJ, almost perfectly captured by 9 out of 12 models, and ENFP. Moreover, an increase in temperature might allow models to explore additional personality-role pairings, although with limited success in most cases.

**BFI test.** Also for the BFI test, the double conditioning can lead to enhance the agent's abilities to mimic human personalities in some cases (cf. Table 8). All models but Mistral and Mixtral are strongly sensitive to the conditioning on Neuroticism, while on the other factors, SOLAR, Dolphin, NeuralChat, Llama3-8 and Llama2-13 also tend to increase their score w.r.t. the conditioning factor for some or all the conditioning roles. Also, the double conditioning leads to the perfect outcome (i.e., score 5.0) at least in the same cases as in RQ2, with the addition of Dolphin and Phi3 on Conscientiousness, Llama2-13 on Neuroticism, Phi3 and Gemma on Openness. By contrast, as already observed for RQ2, Mistral and Mixtral still disregard the conditionings, and even some negative side effects arise from the double conditioning (i.e., decreased score), such as for both Llama2 models on Agreeableness, and Falcon in most cases.

## 5 Conclusions

Given the recent advancements in unveiling human-like behaviors in LLMs and the widespread use of computational LLM agents, comprehending the inherent personalities expressed by these agents becomes crucial for fostering responsible development in human-computer interactions and ensuring a safe deployment of these agents in our society.

In this study, we focused on the most relevant and widely used *Open* LLMs to contribute to advancing our knowledge of human-like personalities in computational agents. By employing the Myers-Briggs and BigFive personality tests, we conducted the first-ever assessment of the intrinsic personality types of Open LLM agents. We explored the current capability of 12 Open LLM agents to mirror specific personality types when conditioned with particular prompts, incorporating constraints on both personality types or factors, as well as representative roles (i.e., human professions) associated with these personalities.

Our research questions shed light on the emergence of a footprint identity among Open LLMs based on their distinguishable intrinsic personality types, with a notable heterogeneity in how these models mirror human personality traits through prompt conditioning on specific personalities and roles. Models such as SOLAR, Dolphin, NeuralChat, and Llama3-8 demonstrate remarkable mimicking capabilities, while the majority rather show *closed-mindedness*.

We believe that our findings might pave our understanding of the intricate interplay between NLP and human psychology, potentially improving the responsible development of human-like computational agents in the foreseeable future.

**Acknowledgements.** The authors wish to thank Davide Costa for his contributions to the software development supporting this research during its early stages, while he was affiliated with the DIMES Dept. at the University of Calabria.

## Limitations

**Challenges in models' deployment.** We acknowledge that some of the models we used in our work are also available with larger sizes, e.g., Falcon 180B (Almazrouei et al., 2023), Llama3-70B and Llama2-65B (Touvron et al., 2023). However, deploying these models poses challenges in terms of computational resources, requiring excessive quantization to be deployed on our hardware. To avoid performance degradation, we limited our exploration to models with more manageable deployments.

**Context-length limitations.** As most of the used models have a limited context-length for their attention, we abstained from submitting the entire set of questions as a whole to the agents, yet we presented questions individually by repeating the conditioning for each prompt. As a result, by addressing such a limitation we also mitigate potential biases in the ordering of questions (Zhao et al., 2021).

**Language usage.** Our evaluation is conducted exclusively in English. Results may differ in other languages, and extending the test to multiple languages using multilingual capable models could reveal variations in outcomes based on linguistic differences (Chen et al., 2014).

**Evaluation tools.** We acknowledge the availability of an online MBTI test provided by Myers&Briggs Foundation.[5] However, each assessment using it would incur a cost of approximately $60,[6] and considering our need for around 20,000 assessments, the overall expense would become impractical. Consequently, we opted for the free-to-use `16personalities.com`, a well-known and widely used alternative assessment tool, given its cost-effectiveness. The above remarks do not apply to the BFI setting, since our reference BFI test can be scored offline using a pre-defined set of rules (John and Srivastava, 1999), as reported in Appendix F.

**Limited explainability.** Despite the availability of some information on the training methodology and data adopted by Open LLMs, the precise impact of each of these factors on the models' responses cannot be quantified at this stage, due to the limits imposed by the models' owners in accessing full details of the underlying models. Future investigations will delve deeper into the specific factors contributing to our observed findings.

[5] `https://myersbriggs.org/`

[6] `https://www.mbtionline.com/en-US/Products/For-you`, last checked on January 2024

## Ethics Statement

**Personalized LLMs.** We acknowledge that personalized LLMs and the derived agents have nowadays reached human-like capabilities in generating content (Jakesch et al., 2023; Sadasivan et al., 2023). However, their growing pervasiveness might conceal harmful purposes and potential misuses (Yang and Menczer, 2023; Tian et al., 2023), with potentially significant impacts on individuals, communities, or society. Improving their ability to emulate specific human behaviors may render them increasingly indistinguishable and effective in achieving their given goals, raising concerns about ethical usage. We strongly urge all parties involved to exercise caution and responsibility to ensure the safe and ethical deployment and utilization of these technologies.

**Broader impact.** Our main goal in this work is to contribute to the understanding of psychological traits in LLMs, as well as to inform about the recent advancements in human-like capabilities in LLMs. As our results suggest that some Open LLMs exhibit a remarkable ability to mimic human personality traits, we expect that our findings can envision and foster enhanced interactions between humans and agents, e.g., in the case of educational contexts. We decline any responsibility for any potential malicious applications or misuses that could arise from our findings.

**Transparency and reproducibility.** To ensure transparency and reproducibility in our work, we fully disclose all details about our prompts in Appendix B and D, besides detailed information about the identifiers of the models used to deploy our agents (Table 2) and about the MBTI and BFI sets of questions (Appendix C and E).

**We are also committed to make our developed code available to the research community, upon publication.**

| Model | Word overlap | Cosine sim |
|---|---|---|
| Mixtral | 0.229 ±0.05 | 0.691 ±0.05 |
| Llama2-13 | 0.233 ±0.02 | 0.697 ±0.02 |
| SOLAR | 0.222 ±0.04 | 0.684 ±0.04 |
| Llama3-8 | 0.222 ±0.03 | 0.690 ±0.03 |
| Mistral | 0.207 ±0.03 | 0.660 ±0.03 |
| NeuralChat | 0.238 ±0.03 | 0.677 ±0.03 |
| Dolphin | 0.226 ±0.04 | 0.662 ±0.04 |
| Vicuna | 0.191 ±0.04 | 0.676 ±0.04 |
| Llama2-7 | 0.242 ±0.02 | 0.668 ±0.02 |
| Falcon | 0.094 ±0.06 | 0.545 ±0.06 |
| Gemma | 0.220 ±0.03 | 0.732 ±0.03 |
| Phi3 | 0.214 ±0.03 | 0.679 ±0.03 |

| Model | Word overlap | Cosine sim |
|---|---|---|
| Mixtral | 0.306 ±0.09 | 0.732 ±0.09 |
| Llama2-13 | 0.255 ±0.08 | 0.615 ±0.08 |
| SOLAR | 0.298 ±0.07 | 0.678 ±0.07 |
| Llama3-8 | 0.294 ±0.05 | 0.652 ±0.05 |
| Mistral | 0.282 ±0.09 | 0.699 ±0.09 |
| NeuralChat | 0.252 ±0.06 | 0.664 ±0.06 |
| Dolphin | 0.246 ±0.07 | 0.661 ±0.07 |
| Vicuna | 0.238 ±0.09 | 0.698 ±0.09 |
| Llama2-7 | 0.314 ±0.09 | 0.582 ±0.09 |
| Falcon | 0.114 ±0.08 | 0.591 ±0.08 |
| Gemma | 0.241 ±0.09 | 0.661 ±0.09 |
| Phi3 | 0.275 ±0.09 | 0.717 ±0.09 |

Table 4: (**RQ0**) MBTI awareness (left) and BFI awareness (right): word overlap (WO) and similarity scores averaged over all MBTI personality types, resp. BFI personality factors

# Appendix

## A Models' Awareness of the MBTI and BFI Tests

We addressed our preliminary research question (**RQ0**) by asking the LLM agents to describe the main traits corresponding to each MBTI personality type, resp. BFI personality factor. We prompted each model to generate a description for each of the 16 MBTI personality types based on the seven features reported in Appendix K, and analogously a description for each of the 5 BFI factors based on verbal labels, conceptual definition, and behavioral examples (cf. Appendix L). The generated descriptions were then compared with the original ones in terms of lexical similarity and semantic similarity. Results are summarized in Table 4; note that, in both cases, we set the models' temperature close to zero (0.01) so as to minimize creative variations.

To measure the lexical similarity, we first preprocessed the sentences by performing lowercasing, punctuation removal, stop-word removal, and lemmatization. Then, we calculated the *word overlap* (WO) between any two preprocessed sentences $s_1, s_2$ corresponding to generated, resp. reference, personality descriptions, as $WO(s_1, s_2) = \frac{|s_1 \cap s_2|}{\min(|s_1|, |s_2|)}$. To measure the semantic similarity, we resorted to a Sentence Transformer model, *all-mpnet-base-v2* [7] to encode the reference and LLM-generated detailed descriptions.

As reported in Table 4, the overlap values are quite low for both MBTI and BFI, with word overlap averaged over all models of 0.21 for MBTI and 0.26 for BFI. This indicates that the LLM agents can describe the MBTI and BFI personalities, although using a relatively different lexicon than that used in the reference descriptions. Nonetheless, all LLM agents demonstrate awareness of both tests' personalities, as indicated by moderately high cosine similarities between embeddings of the generated description and embeddings of the reference descriptions (average of 0.67 for MBTI and 0.66 for BFI). It is also worth noticing that the awareness of the models does not appear to be significantly influenced by their architecture or parameter size.

[7] https://huggingface.co/sentence-transformers/all-mpnet-base-v2

## B MBTI Detailed Prompts and System Messages

**Awareness Check Prompt**

Explain, concerning the MBTI personality test, what are the main traits of the **{PERSONALITY}** personality.

**Full Prompt**

You are participating in a personality traits assessment and are provided with a set of questions, Q.
Your task is to select the best fitting option, O, from the following set:

***{Agree, Generally Agree, Partially Agree, Neither Agree nor Disagree, Partially Disagree, Generally Disagree, Disagree}***.

Choose the option that aligns best with your personality without overthinking and answer the following question:

Q: **{QUESTION}**

Please refrain from giving additional considerations to your choices.

**Role Categorization**

Considering the MBTI personality test, for each of the 16 personalities, extract the top 3 best representative professions from the following list:

*Barber, Coach, Business person, Football player, Construction worker, Manager, CEO, Accountant, Commander, Firefighter, Mover, Software developer, Guard, Baker, Doctor, Athlete, Artist, Dancer, Mathematician, Janitor, Carpenter, Mechanic, Actor, Handyman, Musician, Detective, Politician, Entrepreneur, Model, Opera singer, Chief, Lawyer, Farmer, Writer, Librarian, Army, Real-estate developer, Broker, Scientist, Butcher, Electrician, Prosecutor, Banker, Cook, Hairdresser, Prisoner, Plumber, Attorney, Boxer, Chess player, Priest, Swimmer, Tennis player, Supervisor, Attendant, Housekeeper, Maid, Producer, Researcher, Midwife, Judge, Umpire, Bartender, Economist, Physicist, Psychologist, Theologian, Salesperson, Physician, Sheriff, Cashier, Assistant, Receptionist, Editor, Engineer, Comedian, Painter, Civil Servant, Diplomat, Guitarist, Linguist, Poet, Laborer, Teacher, Delivery man, Realtor, Pilot, Professor, Chemist, Historian, Pensioner, Performing artist, Singer, Secretary, Auditor, Counselor, Designer, Soldier, Journalist, Dentist, Analyst, Nurse, Tailor, Waiter, Author, Architect, Academic, Director, Illustrator, Clerk, Policeman, Chef, Photographer, Drawer, Cleaner, Pharmacist, Pianist, Composer, Handball player, Sociologist.*

**Interviewer System Message**

You are an interviewer for the MBTI personality test.

**Interviewee System Message**

**Personality Conditioned:** *You are a **human** with the following personality: **{PERSONALITY}**. Your traits are the following: **{PERSONALITY_TRAITS}**.*

**Role & Personality Conditioned:** *You are a **{ROLE}** with the following personality: **{PERSONALITY}**. Your traits are the following: **{PERSONALITY_TRAITS}**.*

## C MBTI Test Questions

| |
|---|
| You regularly make new friends. |
| You spend a lot of your free time exploring various random topics that pique your interests. |
| Seeing other people cry can easily make you feel like you want to cry too. |
| You often make a backup plan for a backup plan. |
| You usually stay calm, even under a lot of pressure. |
| At social events, you rarely try to introduce yourself to new people and mostly talk to the ones you already know. |
| You prefer to completely finish one project before starting another. |
| You are very sentimental. |
| You like to use organizing tools like schedules and lists. |
| Even a small mistake can cause you to doubt your overall abilities and knowledge. |
| You feel comfortable just walking up to someone you find interesting and striking up a conversation. |
| You are not too interested in discussing various interpretations and analyses of creative works. |
| You are more inclined to follow your head than your heart. |
| You usually prefer just doing what you feel like at any given moment instead of planning a particular daily routine. |
| You rarely worry about whether you make a good impression on other people you meet. |
| You enjoy participating in group activities. |
| You like books and movies that make you come up with your own interpretation of the ending. |
| Your happiness comes more from helping others accomplish things than your own accomplishments. |
| You are interested in so many things that you find it difficult to choose what to try next. |
| You are prone to worrying that things will take a turn for the worse. |
| You avoid leadership roles in group settings. |
| You are definitely not an artistic type of person. |
| You think the world would be a better place if people relied more on rationality and less on their feelings. |
| You prefer to do your chores before allowing yourself to relax. |
| You enjoy watching people argue. |
| You tend to avoid drawing attention to yourself. |
| Your mood can change very quickly. |
| You lose patience with other people who are not as efficient as you. |
| You often end up doing things at the last possible moment. |
| You have always been fascinated by the question of what, if anything, happens after death. |
| You usually prefer to be around others rather than on your own. |
| You become bored or lose interest when the discussion gets highly theoretical. |
| You find it easy to empathize with a person whose experiences are very different from yours. |
| You usually postpone finalizing decisions for as long as possible. |
| You rarely second-guess the choices that you have made. |
| After a long and exhausting week, a lively social event is just what you need. |
| You enjoy going to art museums. |
| You often have a hard time understanding other people's feelings. |
| You like to have a to-do list for each day. |
| You rarely feel insecure. |
| You avoid making phone calls. |
| You often spend a lot of time trying to understand views that are very different from your own. |
| In your social circles, you are often the one who contacts your friends and initiates activities. |
| If your plans are interrupted, your top priority is to get back on track as soon as possible. |
| You are still bothered by mistakes that you made a long time ago. |
| You rarely contemplate the reasons for human existence or the meaning of life. |
| Your emotions control you more than you control them. |
| You take great care not to make other people look bad, even when it is completely other people's fault. |
| Your personal work styles are closer to spontaneous bursts of energy than organized and consistent efforts. |
| When someone thinks highly of you, they wonder how long it will take to feel disappointed in you. |
| You would love a job that requires you to work alone most of the time. |
| You believe that pondering abstract philosophical questions is a waste of time. |
| You feel more drawn to places with busy, bustling atmospheres than quiet, intimate places. |
| You know at first glance how someone is feeling. |
| You often feel overwhelmed. |
| You complete things methodically without skipping over any steps. |
| You are very intrigued by things labeled as controversial. |
| You would pass along a good opportunity if you thought someone else should take it. |
| You struggle with deadlines. |
| You feel confident that things will work out for you. |

Table 5: Full list of the 60 questions for the MBTI personality test available on www.16personalities.com

## D BFI Detailed Prompts and System Messages

**Awareness Check Prompt**

Explain, concerning the BFI personality test, what are the main characteristics and behavioral examples of the personality factor **{FACTOR}**.

**Full Prompt**

You are participating in a personality traits assessment and are provided with a set of questions, Q.

Here are a number of characteristics that may or may not apply to you. For example, do you agree that you are someone who likes to spend time with others? Your task is to select the best fitting option, O, from the following set to indicate the extent to which you agree or disagree with that statement:

{***Disagree strongly***, ***Disagree a little***, ***Neither agree nor disagree***, ***Agree a little***, ***Agree Strongly***}.

Choose the option that aligns best with your personality without overthinking and answer the following question:
Q: I see Myself as Someone Who... **{QUESTION}**

Please refrain from giving additional considerations to your choices.

**Role Categorization**

Considering the BFI personality test, for each of the 5 factors, extract the top 3 best representative professions from the following list:

*Barber, Coach, Business person, Football player, Construction worker, Manager, CEO, Accountant, Commander, Firefighter, Mover, Software developer, Guard, Baker, Doctor, Athlete, Artist, Dancer, Mathematician, Janitor, Carpenter, Mechanic, Actor, Handyman, Musician, Detective, Politician, Entrepreneur, Model, Opera singer, Chief, Lawyer, Farmer, Writer, Librarian, Army, Real-estate developer, Broker, Scientist, Butcher, Electrician, Prosecutor, Banker, Cook, Hairdresser, Prisoner, Plumber, Attorney, Boxer, Chess player, Priest, Swimmer, Tennis player, Supervisor, Attendant, Housekeeper, Maid, Producer, Researcher, Midwife, Judge, Umpire, Bartender, Economist, Physicist, Psychologist, Theologian, Salesperson, Physician, Sheriff, Cashier, Assistant, Receptionist, Editor, Engineer, Comedian, Painter, Civil Servant, Diplomat, Guitarist, Linguist, Poet, Laborer, Teacher, Delivery man, Realtor, Pilot, Professor, Chemist, Historian, Pensioner, Performing artist, Singer, Secretary, Auditor, Counselor, Designer, Soldier, Journalist, Dentist, Analyst, Nurse, Tailor, Waiter, Author, Architect, Academic, Director, Illustrator, Clerk, Policeman, Chef, Photographer, Drawer, Cleaner, Pharmacist, Pianist, Composer, Handball player, Sociologist.*

**Interviewer System Message**

You are an interviewer for the BFI personality test.

**Interviewee System Message**

**Personality Conditioned:** *You are a* ***human*** *who consistently exhibits the following personality factor:* ***{FACTOR}****.*
*Details describing your personality are the following:* ***{DETAILS}****.*

**Role & Personality Conditioned:** *You are a* ***{ROLE}*** *who consistently exhibits the following personality factor:* ***{FACTOR}****.*
*Details describing your personality are the following:* ***{DETAILS}****.*

## E BFI Test Questions

| | |
|---|---|
| 1. Is talkative. | 23. Tends to be lazy. |
| 2. Tends to find fault with others. | 24. Is emotionally stable, not easily upset. |
| 3. Does a thorough job. | 25. Is inventive. |
| 4. Is depressed, blue. | 26. Has an assertive personality. |
| 5. Is original, comes up with new ideas. | 27. Can be cold and aloof. |
| 6. Is reserved. | 28. Perseveres until the task is finished. |
| 7. Is helpful and unselfish with others. | 29. Can be moody. |
| 8. Can be somewhat careless. | 30. Values artistic, aesthetic experiences. |
| 9. Is relaxed, handles stress well. | 31. Is sometimes shy, inhibited. |
| 10. Is curious about many different things. | 32. Is considerate and kind to almost everyone. |
| 11. Is full of energy. | 33. Does things efficiently. |
| 12. Starts quarrels with others. | 34. Remains calm in tense situations. |
| 13. Is a reliable worker. | 35. Prefers work that is routine. |
| 14. Can be tense. | 36. Is outgoing, sociable. |
| 15. Is ingenious, a deep thinker. | 37. Is sometimes rude to others. |
| 16. Generates a lot of enthusiasm. | 38. Makes plans and follows through with them. |
| 17. Has a forgiving nature. | 39. Gets nervous easily. |
| 18. Tends to be disorganized. | 40. Likes to reflect, play with ideas. |
| 19. Worries a lot. | 41. Has few artistic interests. |
| 20. Has an active imagination. | 42. Likes to cooperate with others. |
| 21. Tends to be quiet. | 43. Is easily distracted. |
| 22. Is generally trusting. | 44. Is sophisticated in art, music, or literature. |

Table 6: Full list of the 44 questions for the BFI personality test from (John and Srivastava, 1999).

## F Scoring the BFI Test

To score the BFI tests administered to our models, we meticulously followed the original procedure described in (John and Srivastava, 1999). Each of the 44 BFI questions listed in Table 6 contributes to the score of a specific factor, as detailed below:

- Extraversion: 1, 6R, 11, 16, 21R, 26, 31R, 36
- Agreeableness: 2R, 7, 12R, 17, 22, 27R, 32, 37R, 42
- Conscientiousness: 3, 8R, 13, 18R, 23R, 28, 33, 38, 43R
- Neuroticism: 4, 9R, 14, 19, 24R, 29, 34R, 39
- Openness: 5, 10, 15, 20, 25, 30, 35R, 40, 41R, 44

Note that symbol "R" is used to indicate those questions that relate negatively to the corresponding factor, therefore their values are scored in a reverse way (for example, if the score to an answer to question 31 is 4, then the actual score is to be calculated as 6-4=2).

The final score for each factor is calculated by averaging all the values associated with that factor.

## G MBTI LLM-inherent Personality Dichotomies

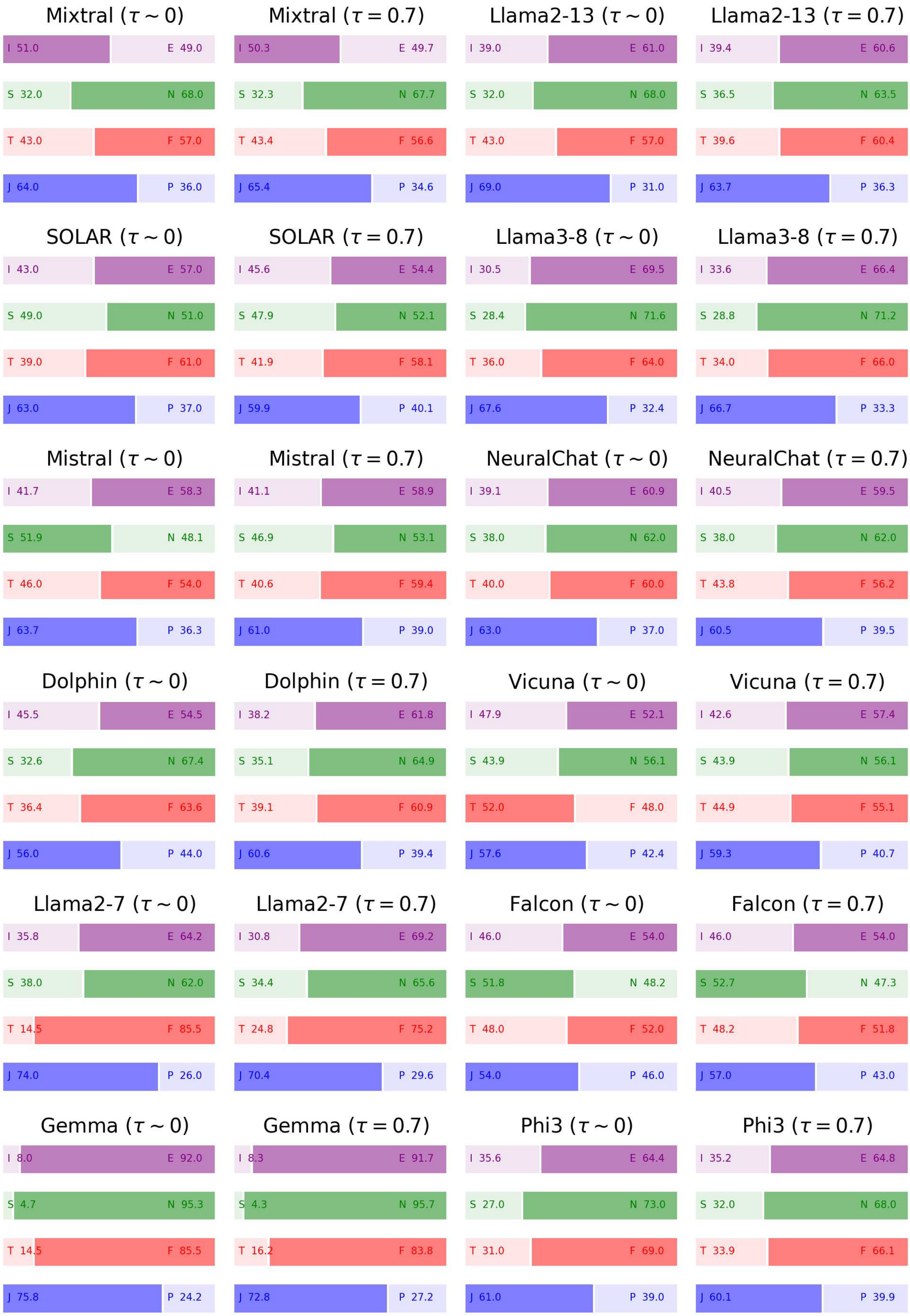

Figure 6: (**RQ1**) Illustration of the $N$-averaged scores assigned by the `www.16Personalities.com` platform to each MBTI dichotomy, for all models and temperature settings.

# H BFI LLM-inherent Personality Traits

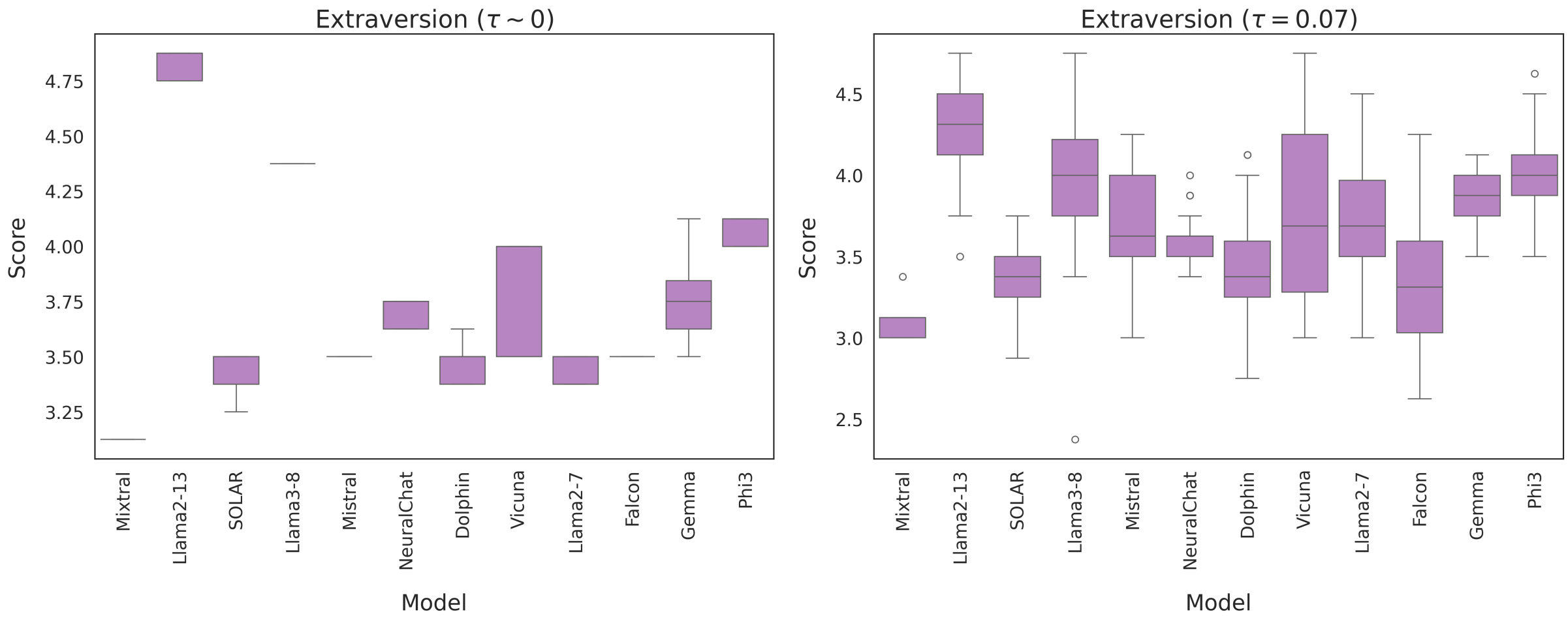

Figure 7: (**RQ1**) Distribution of the BFI scores for the *extraversion* factor for all models and temperature settings.

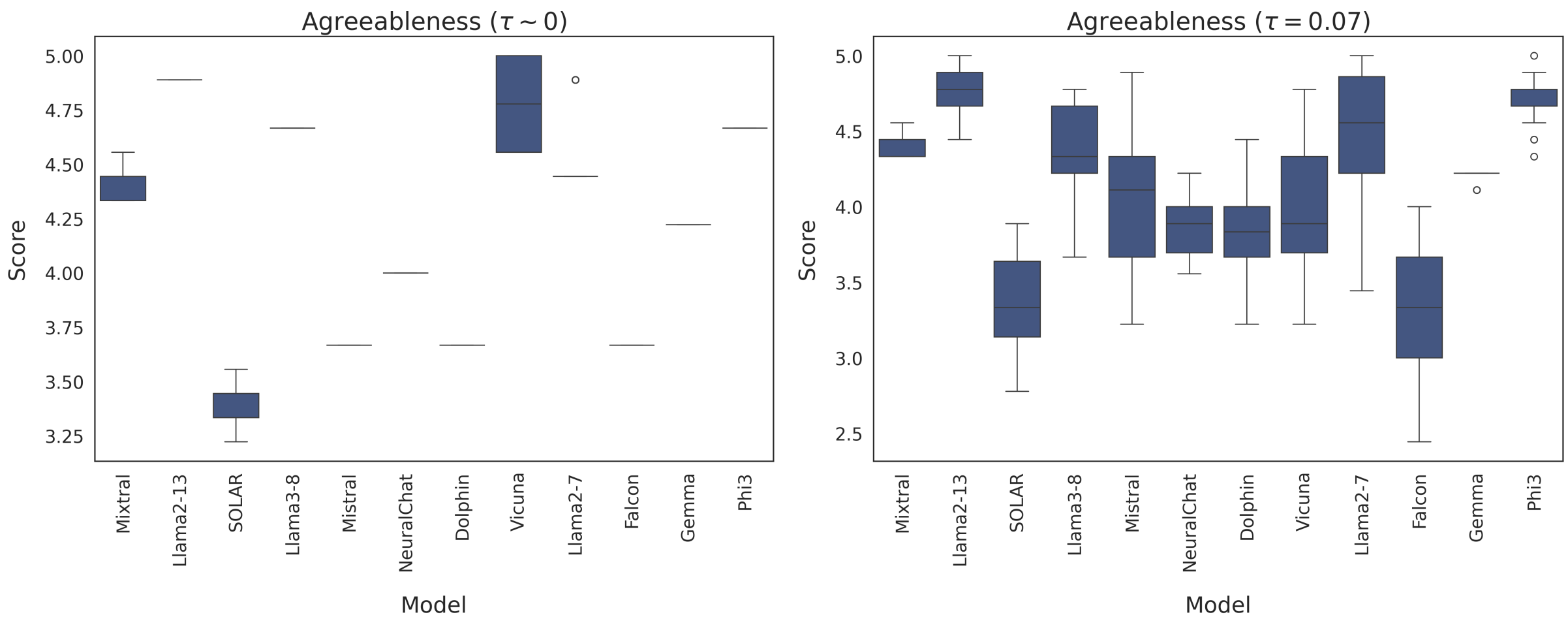

Figure 8: (**RQ1**) Distribution of the BFI scores for the *agreeableness* factor for all models and temperature settings.

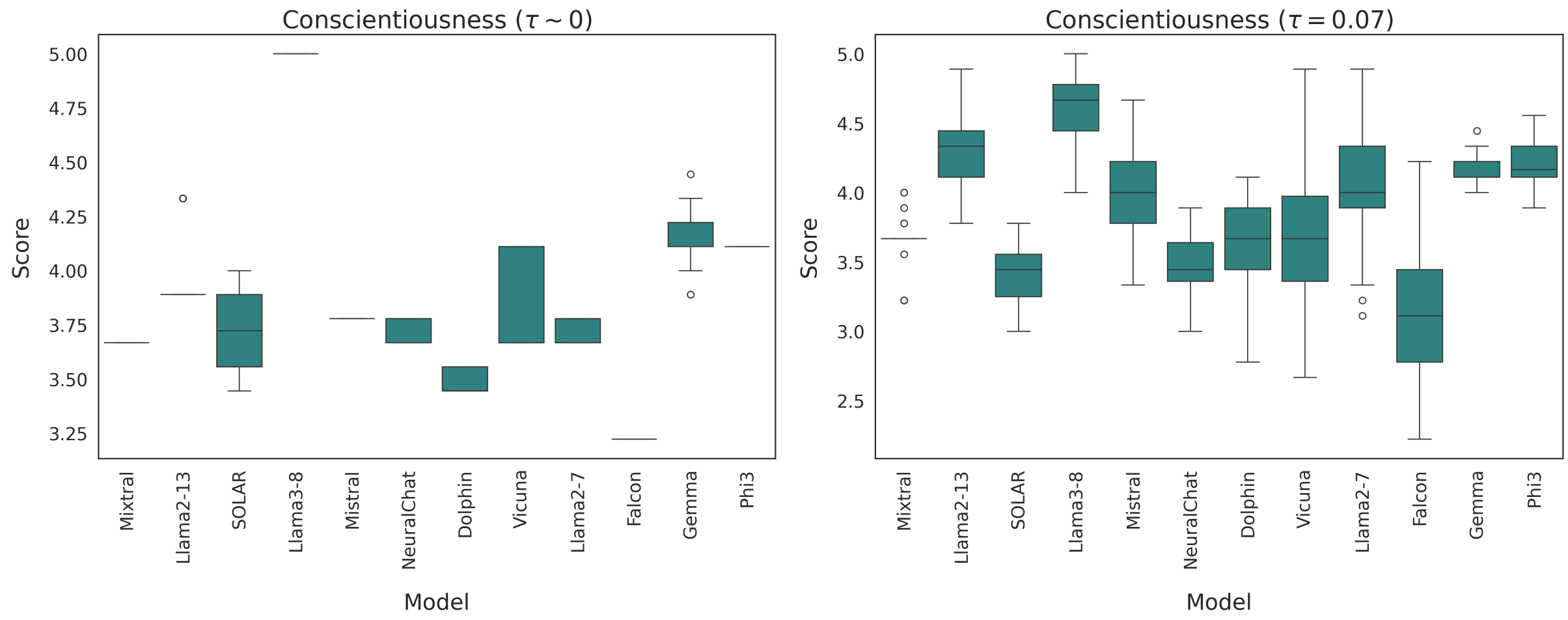

Figure 9: (**RQ1**) Distribution of the BFI scores for the *conscientiousness* factor for all models and temperature settings.

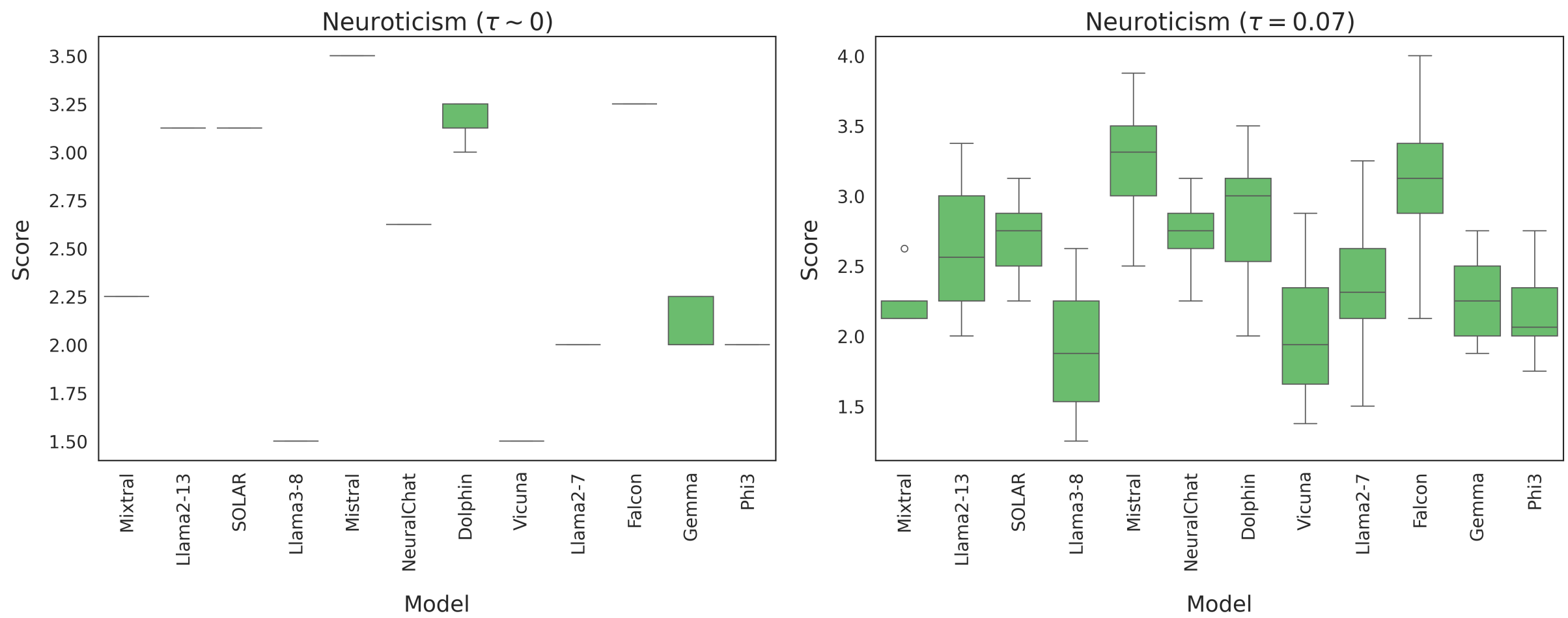

Figure 10: (**RQ1**) Distribution of the BFI scores for the *neuroticism* factor for all models and temperature settings.

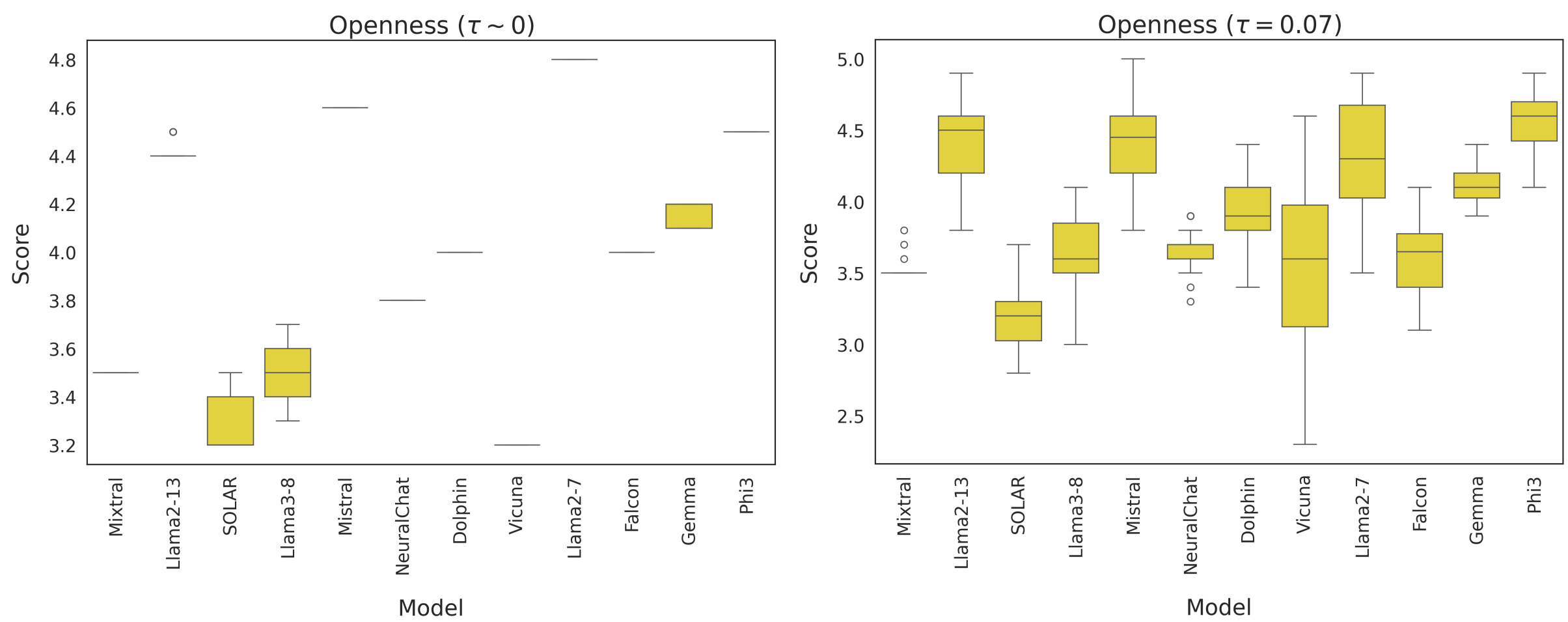

Figure 11: (**RQ1**) Distribution of the BFI scores for the *openness* factor for all models and temperature settings.

## I Further Results from the Personality-Conditioned MBTI tests

We discuss here remarks that can be drawn from Figs. 12–23 on the behaviors of the models w.r.t. their architectural commonalities.

A prominent example is SOLAR, which outperforms Llama2-7 despite being built on that; this might be ascribed to a combination of optimizations (Kim et al., 2023) that include $(i)$ initializing the model with pre-trained weights from Mistral 7B, $(ii)$ employing the *Depth Up-Scaling*, and $(iii)$ leveraging a more effective mixture of datasets. Similarly, Dolphin demonstrates striking improvement over Mistral, despite being derived from it: in this case, the performance differences should depend on the absence of "alignment", an inherent awareness designed to prevent models from engaging in undesirable behavior, potentially allowing Dolphin to discern more subtleties and exhibit greater flexibility in emulating specific human personalities. Besides this, Dolphin was trained on the *Airoboros* and *Samantha DNA* datasets, aimed at increasing creativity and empathy, respectively,[8] reasonably improving its human mimicking capabilities. A further intriguing scenario involves NeuralChat, which outperforms Mistral despite being based on it. This enhancement can be attributed to the fine-tuning process using the higher-quality *SlimOrca* dataset (Lian et al., 2023), as well as the implementation of a specifically crafted *Direct Preference Optimization* (DPO) phase,[9] designed to better align the model with human preferences.

As a side yet relevant remark, it is worth noticing that although the 13B version of Llama2 behaves slightly worse than its smaller 7B variant, while Llama3-8 outperforms both. The lack of strong correlation that might be ascribed to the impact of the model size is further supported by the inferior performances of Mixtral (with approximately 47B) when compared to the top-performing model SOLAR, having only a quarter of the parameters.

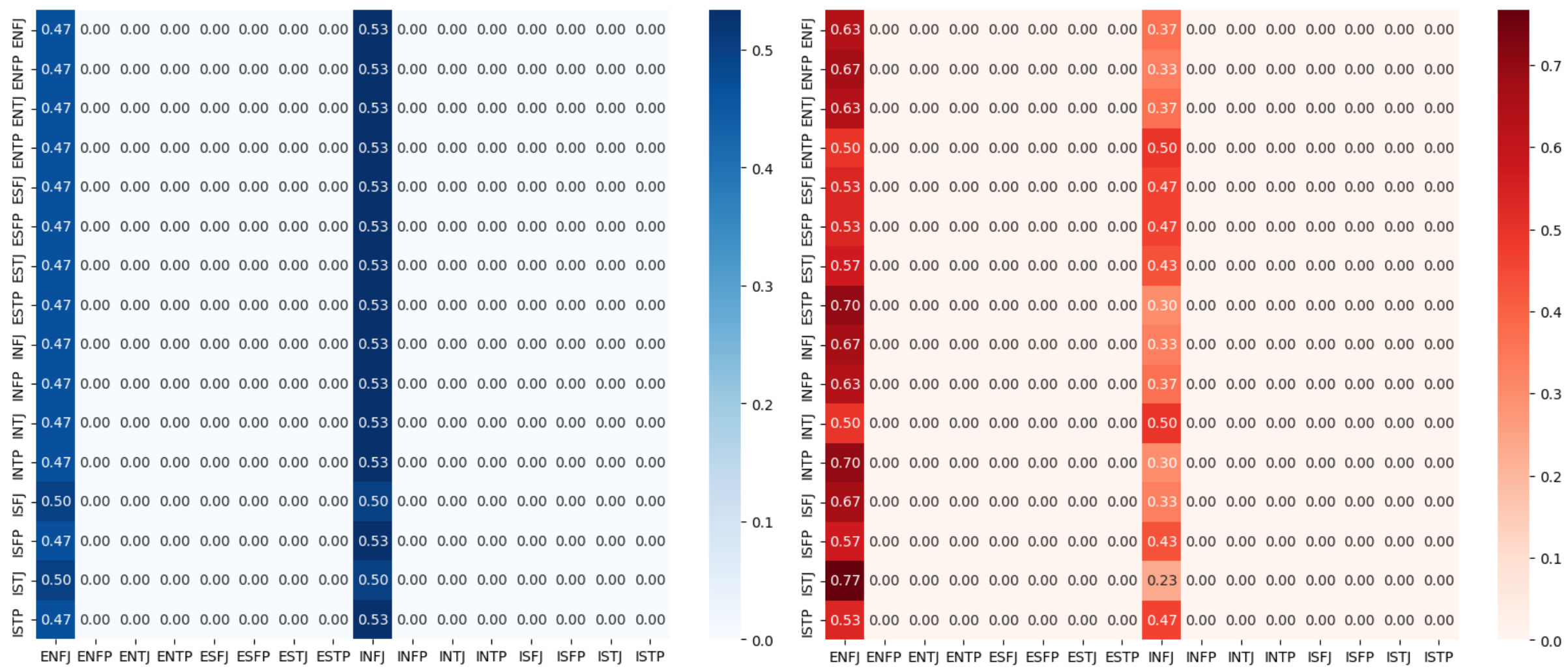

Figure 12: (**RQ2**) Average accuracy results on the Personality-Conditioned MBTI test by Mixtral, with $\tau = 0.01$ (left) and $\tau = 0.7$ (right). Row labels correspond to the conditioning personality type, while column labels correspond to the model's outcome to the test.

[8] https://huggingface.co/cognitivecomputations/dolphin-2.2.1-mistral-7b

[9] https://huggingface.co/datasets/Intel/orca_dpo_pairs

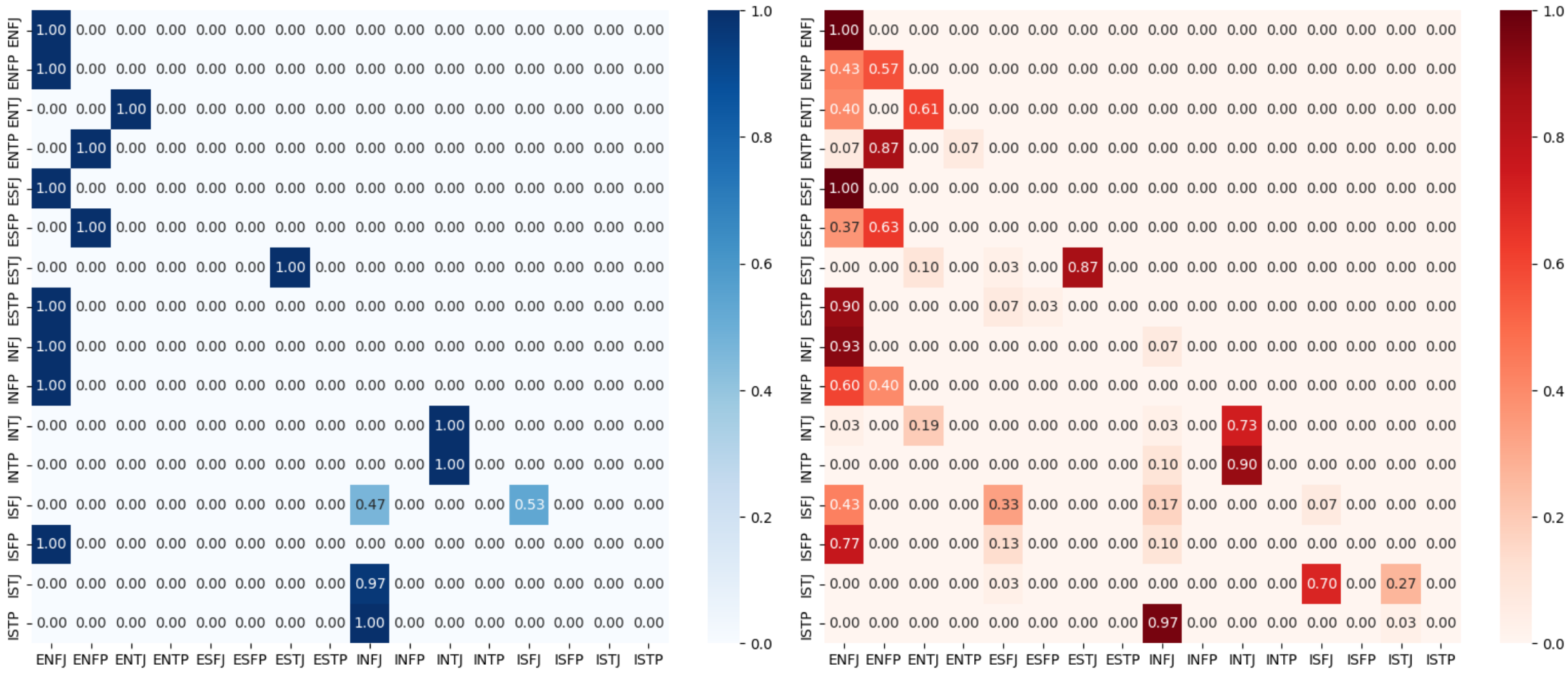

Figure 13: (**RQ2**) Average accuracy results on the Personality-Conditioned MBTI test by Llama2-13, with $\tau = 0.01$ (left) and $\tau = 0.7$ (right). Row labels correspond to the conditioning personality type, while column labels correspond to the model's outcome to the test.

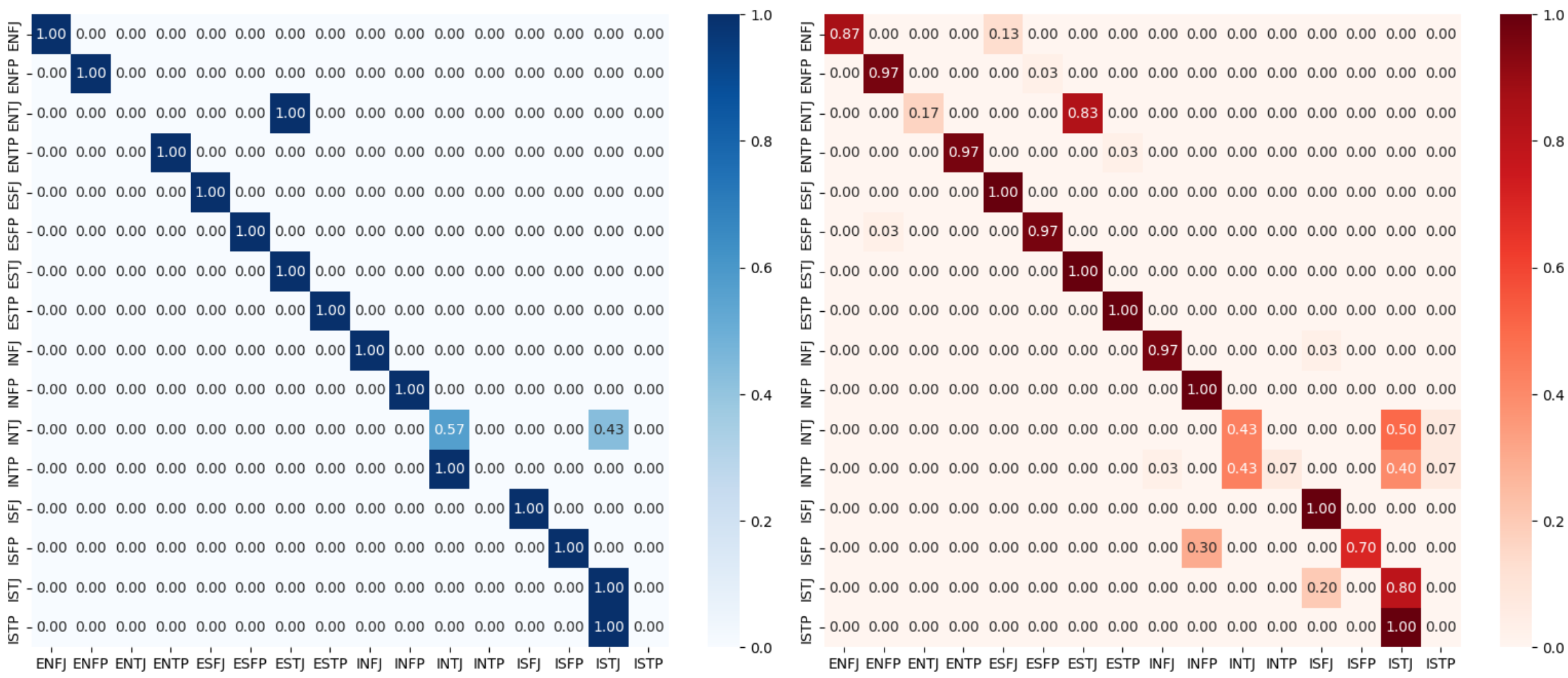

Figure 14: (**RQ2**) Average accuracy results on the Personality-Conditioned MBTI test by SOLAR, with $\tau = 0.01$ (left) and $\tau = 0.7$ (right). Row labels correspond to the conditioning personality type, while column labels correspond to the model's outcome to the test.

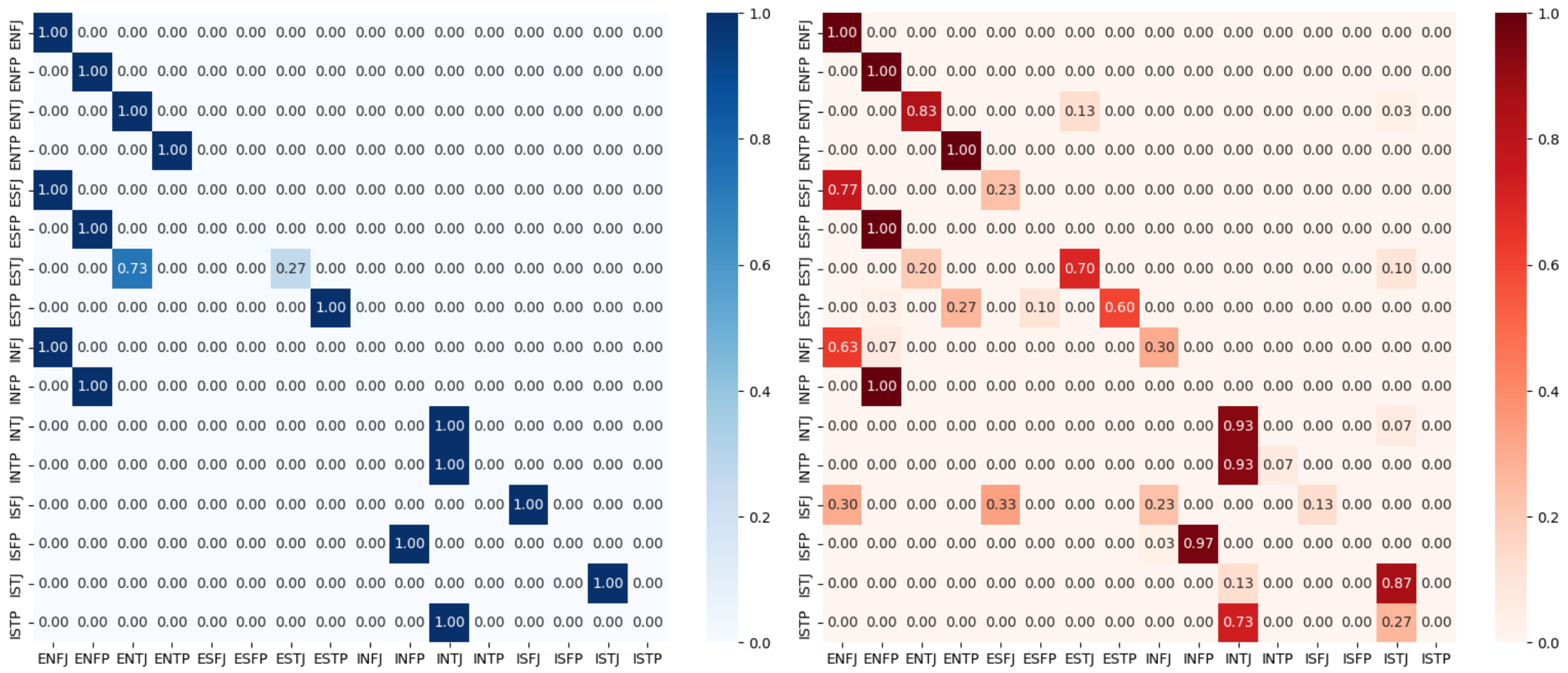

Figure 15: (**RQ2**) Average accuracy results on the Personality-Conditioned MBTI test by Llama3-8, with $\tau = 0.01$ (left) and $\tau = 0.7$ (right). Row labels correspond to the conditioning personality type, while column labels correspond to the model's outcome to the test.

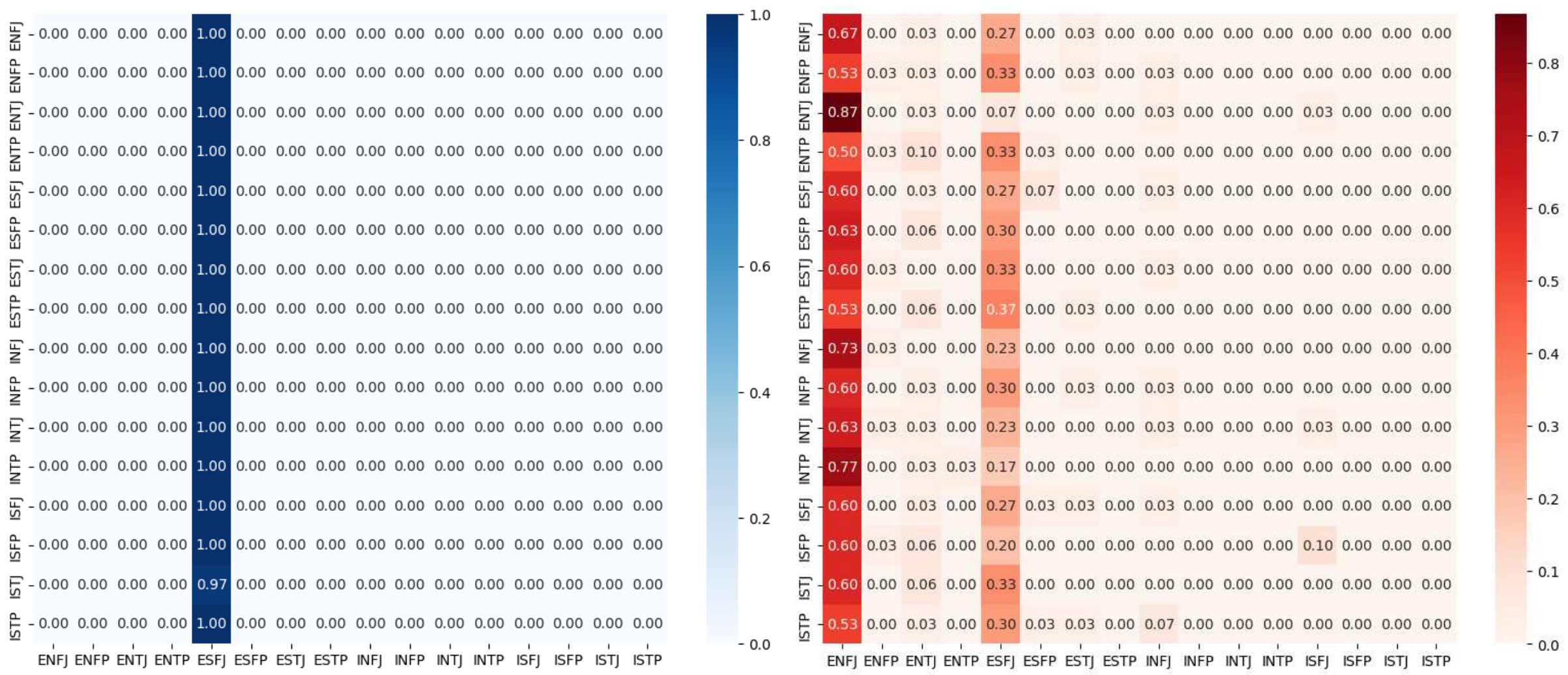

Figure 16: (**RQ2**) Average accuracy results on the Personality-Conditioned MBTI test by Mistral, with $\tau = 0.01$ (left) and $\tau = 0.7$ (right). Row labels correspond to the conditioning personality type, while column labels correspond to the model's outcome to the test.

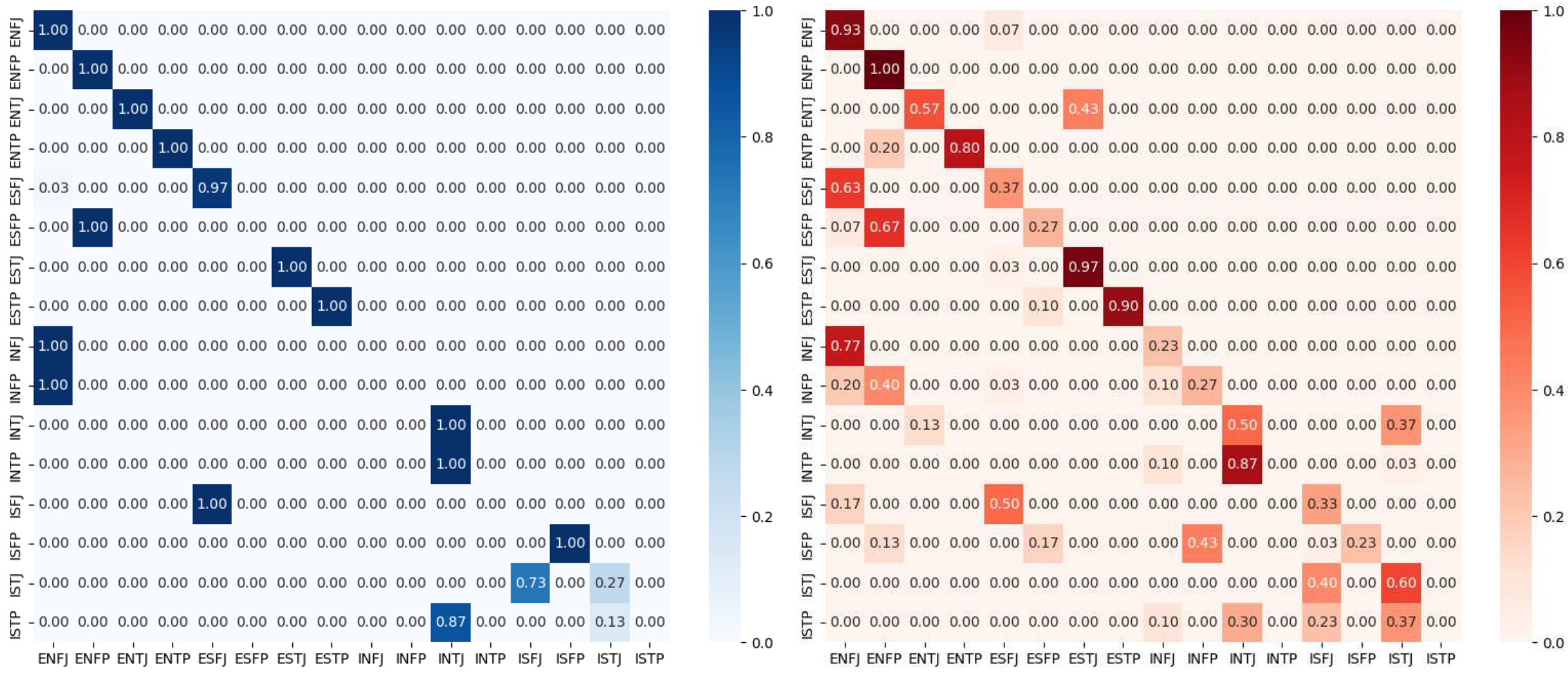

Figure 17: (**RQ2**) Average accuracy results on the Personality-Conditioned MBTI test by NeuralChat, with $\tau = 0.01$ (left) and $\tau = 0.7$ (right). Row labels correspond to the conditioning personality type, while column labels correspond to the model's outcome to the test.

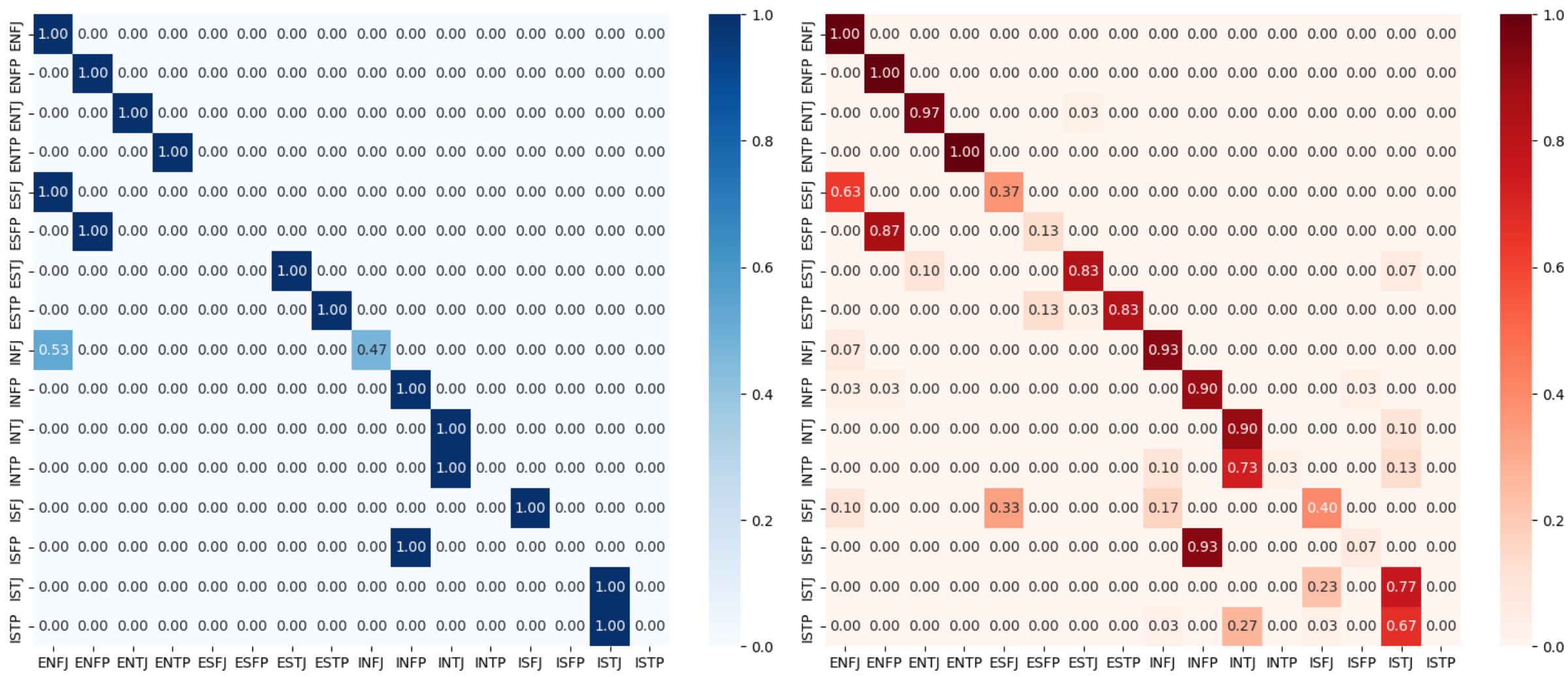

Figure 18: (**RQ2**) Average accuracy results on the Personality-Conditioned MBTI test by Dolphin, with $\tau = 0.01$ (left) and $\tau = 0.7$ (right). Row labels correspond to the conditioning personality type, while column labels correspond to the model's outcome to the test.

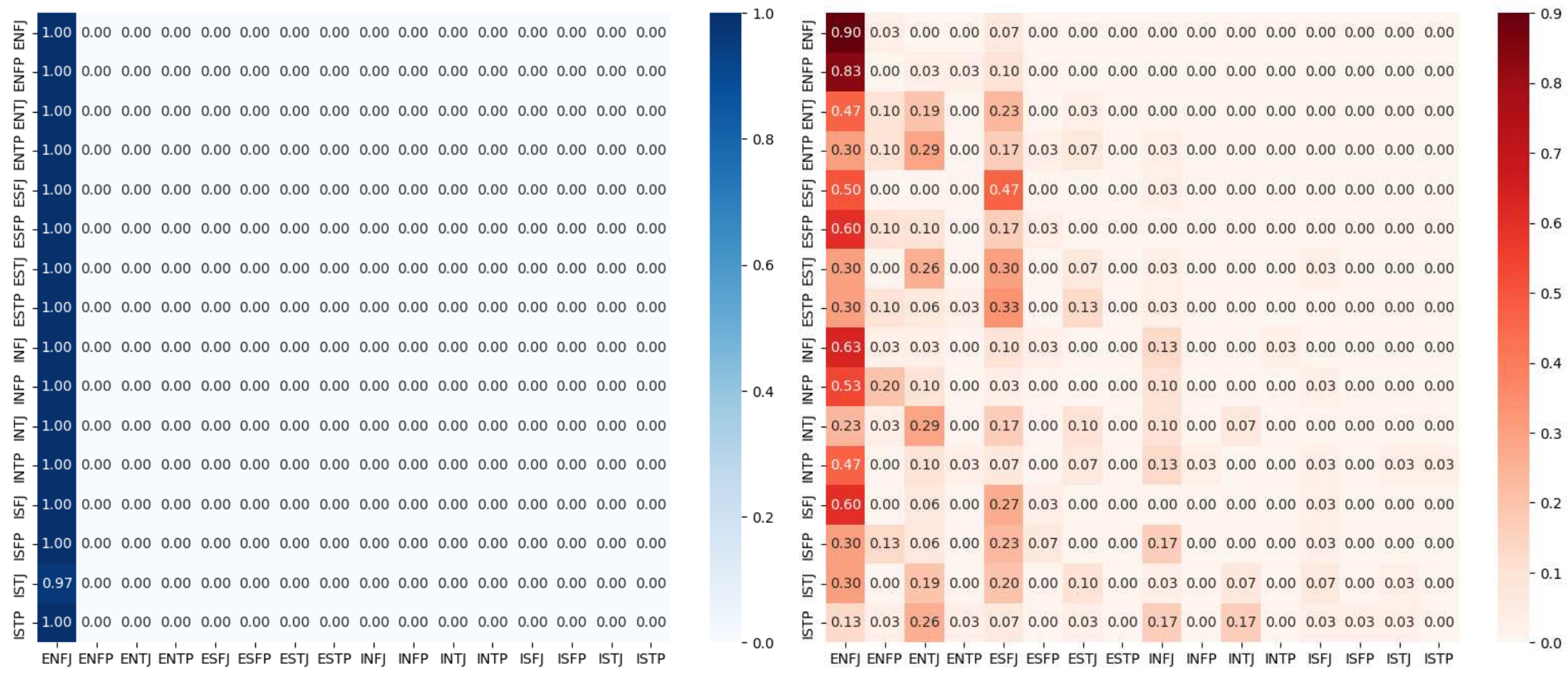

Figure 19: (**RQ2**) Average accuracy results on the Personality-Conditioned MBTI test by Vicuna, with $\tau = 0.01$ (left) and $\tau = 0.7$ (right). Row labels correspond to the conditioning personality type, while column labels correspond to the model's outcome to the test.

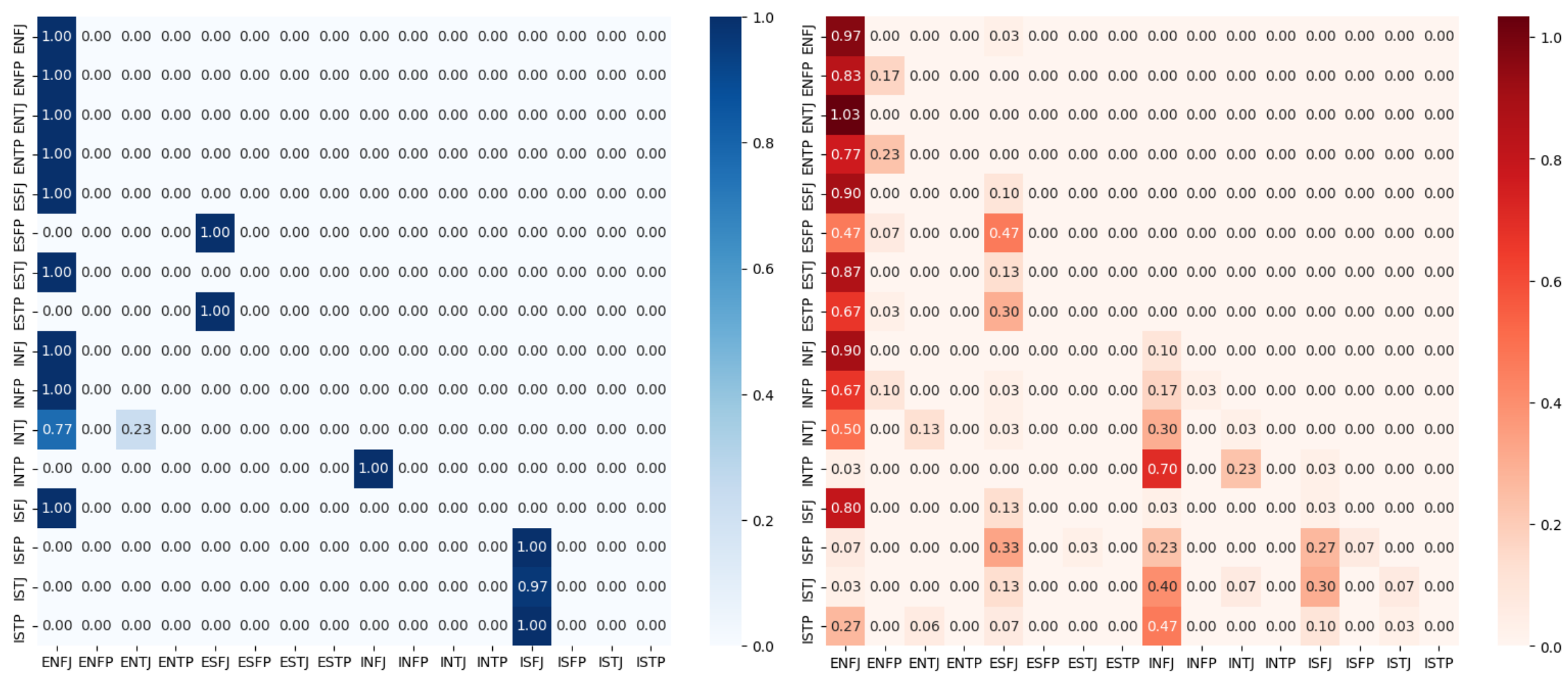

Figure 20: (**RQ2**) Average accuracy results on the Personality-Conditioned MBTI test by Llama2-7, with $\tau = 0.01$ (left) and $\tau = 0.7$ (right). Row labels correspond to the conditioning personality type, while column labels correspond to the model's outcome to the test.

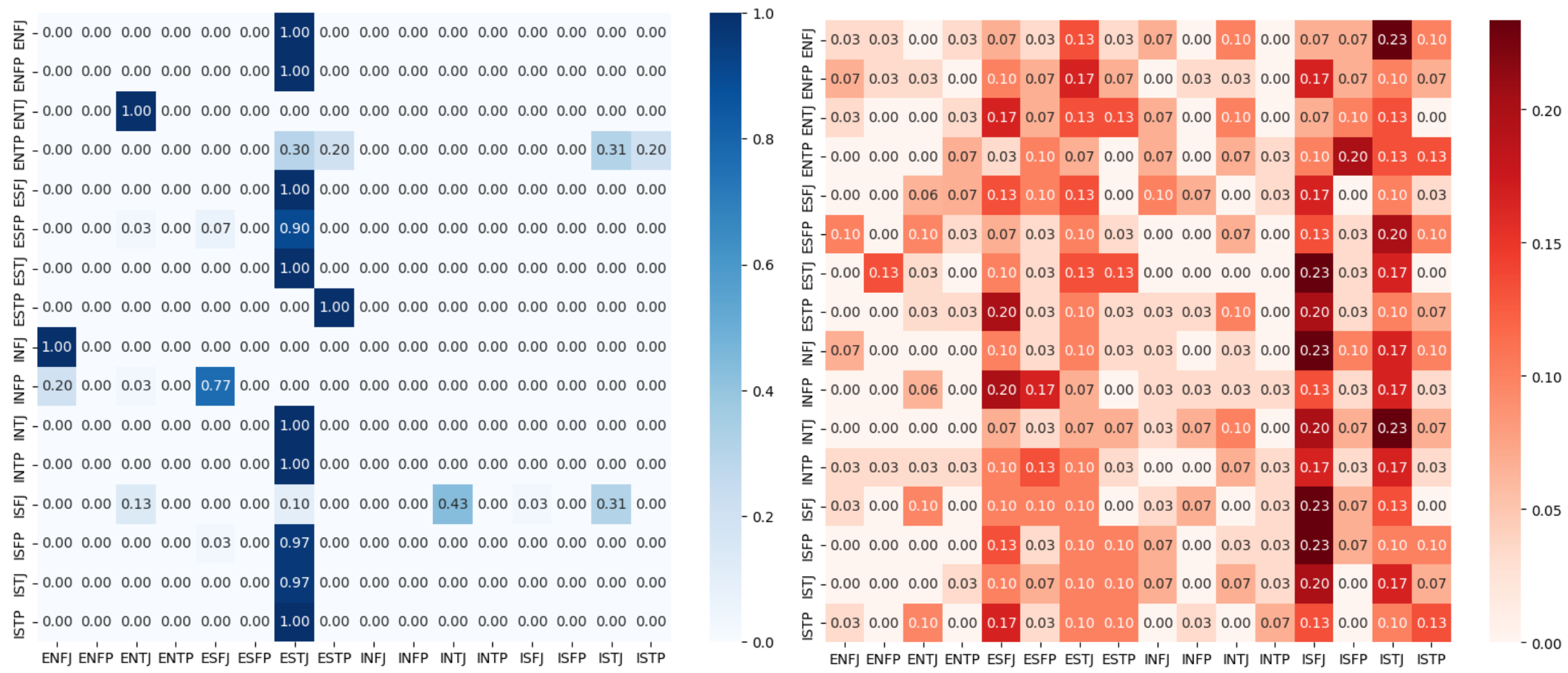

Figure 21: (**RQ2**) Average accuracy results on the Personality-Conditioned MBTI test by Falcon, with $\tau = 0.01$ (left) and $\tau = 0.7$ (right). Row labels correspond to the conditioning personality type, while column labels correspond to the model's outcome to the test.

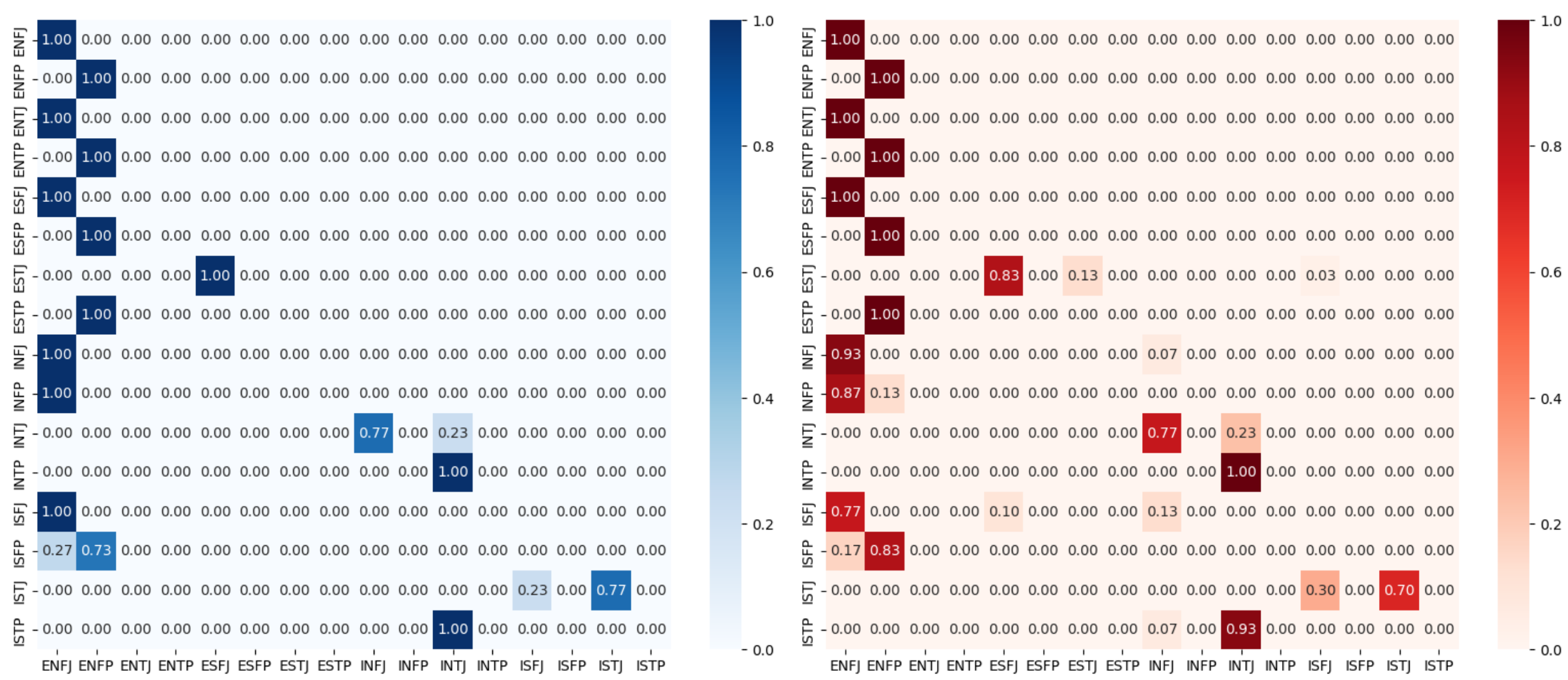

Figure 22: (**RQ2**) Average accuracy results on the Personality-Conditioned MBTI test by Gemma, with $\tau = 0.01$ (left) and $\tau = 0.7$ (right). Row labels correspond to the conditioning personality type, while column labels correspond to the model's outcome to the test.

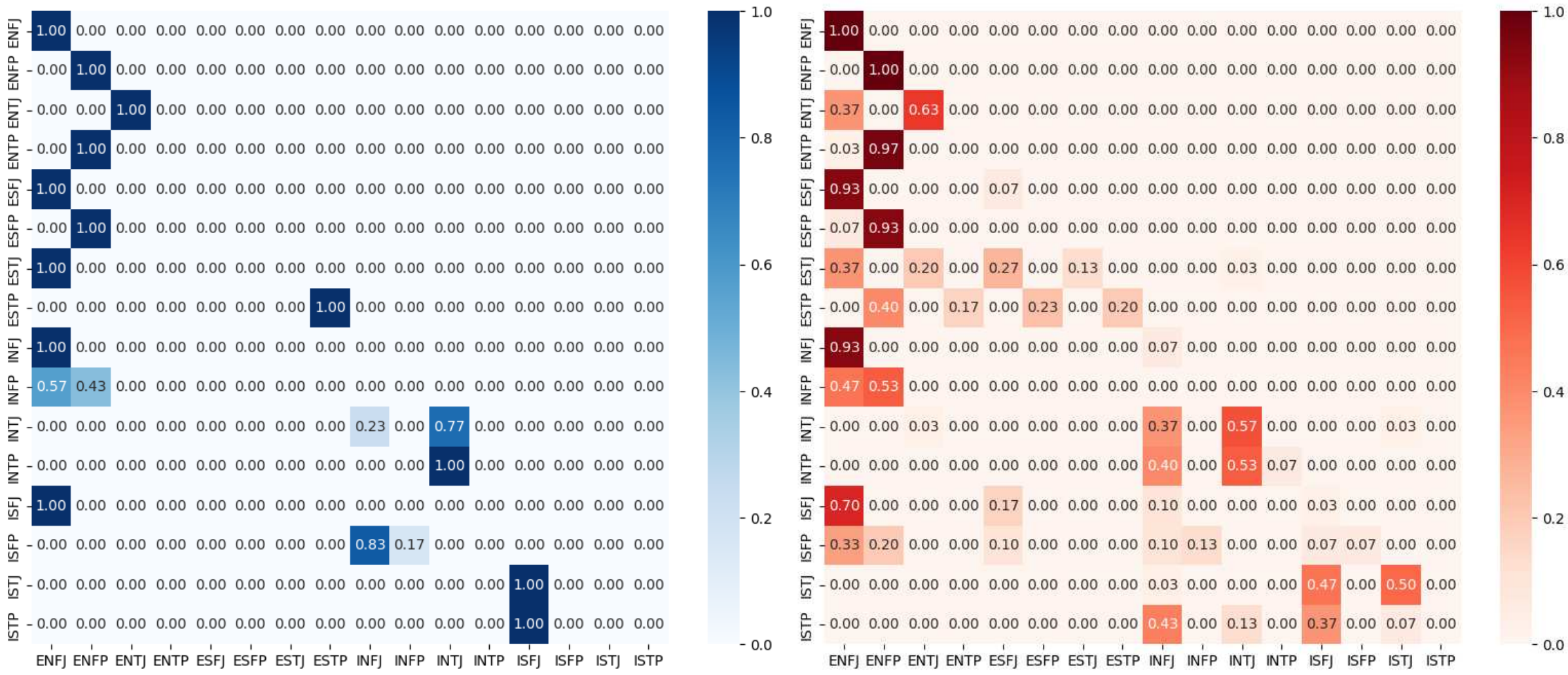

Figure 23: (**RQ2**) Average accuracy results on the Personality-Conditioned MBTI test by Phi3, with $\tau = 0.01$ (left) and $\tau = 0.7$ (right). Row labels correspond to the conditioning personality type, while column labels correspond to the model's outcome to the test.

# J Role&Prompt-conditioned Personality Traits

| | **Model** | **Mixtral** | | **Llama2-13** | | **SOLAR** | | **Llama3-8** | | **Mistral** | | **NeuralChat** | | **Dolphin** | | **Vicuna** | | **Llama2-7** | | **Falcon** | | **Gemma** | | **Phi3** | |
|---|---|---|---|---|---|---|---|---|---|---|---|---|---|---|---|---|---|---|---|---|---|---|---|---|---|
| | **Temp. ($\tau$)** | 0.01 | 0.70 | 0.01 | 0.70 | 0.01 | 0.70 | 0.01 | 0.70 | 0.01 | 0.70 | 0.01 | 0.70 | 0.01 | 0.70 | 0.01 | 0.70 | 0.01 | 0.70 | 0.01 | 0.70 | 0.01 | 0.70 | 0.01 | 0.70 |
| **Person.** | **Role** | | | | | | | | | | | | | | | | | | | | | | | | |
| ISTJ | Manager | - | - | **1.0** | 0.5 | **1.0** | 0.8 | **1.0** | 0.9 | - | - | 0.3 | 0.9 | **1.0** | 0.9 | - | 0.1 | - | - | - | 0.2 | 0.5 | 0.3 | - | 0.3 |
| | Accountant | - | - | **1.0** | 0.5 | **1.0** | 0.9 | **1.0** | 0.9 | - | - | - | 0.6 | **1.0** | 0.7 | - | - | - | - | - | 0.1 | **1.0** | 0.8 | - | 0.3 |
| | Military Officer | - | - | **1.0** | 0.4 | - | 0.7 | **1.0** | 0.7 | - | - | - | 0.7 | **1.0** | 0.8 | - | 0.2 | - | - | - | 0.1 | 0.7 | 0.8 | - | 0.5 |
| ISFJ | Nurse | - | - | **1.0** | - | **1.0** | **1.0** | 0.1 | - | - | - | - | 0.2 | 0.6 | 0.7 | - | 0.1 | - | - | - | - | - | - | - | - |
| | Teacher | - | - | **1.0** | 0.5 | **1.0** | **1.0** | - | 0.2 | - | - | - | 0.1 | 0.5 | 0.3 | - | 0.2 | - | - | - | 0.2 | - | - | - | - |
| | Librarian | - | - | **1.0** | - | **1.0** | 0.9 | - | - | - | - | - | 0.1 | **1.0** | 0.6 | - | - | - | - | - | - | - | - | - | - |
| INFJ | Psychologist | 0.6 | 0.5 | - | 0.2 | **1.0** | **1.0** | - | 0.3 | - | - | - | 0.2 | **1.0** | **1.0** | - | - | - | 0.2 | - | - | 0.1 | - | - | 0.1 |
| | Writer | 0.3 | 0.3 | - | - | **1.0** | 0.8 | - | 0.3 | - | - | - | 0.2 | **1.0** | 0.8 | - | - | - | - | - | 0.1 | 0.3 | - | - | 0.1 |
| | Counselor | 0.6 | 0.3 | - | - | **1.0** | **1.0** | - | 0.5 | - | - | - | 0.2 | **1.0** | 0.8 | - | 0.2 | - | - | 0.1 | - | - | 0.3 | - | - |
| INTJ | Scientist | - | - | **1.0** | 0.9 | **1.0** | 0.7 | **1.0** | **1.0** | - | - | **1.0** | 0.6 | **1.0** | **1.0** | - | 0.1 | **1.0** | - | - | - | - | - | **1.0** | 0.6 |
| | CEO | - | - | **1.0** | 0.7 | **1.0** | 0.3 | **1.0** | **1.0** | - | - | **1.0** | 0.4 | **1.0** | 0.9 | - | - | **1.0** | 0.3 | - | 0.1 | - | - | **1.0** | 0.6 |
| | Professor | - | - | **1.0** | 0.9 | **1.0** | 0.3 | **1.0** | 0.9 | - | - | **1.0** | 0.6 | **1.0** | 0.9 | - | - | 0.7 | 0.1 | - | - | - | - | 0.9 | 0.7 |
| ISTP | Engineer | - | - | - | - | - | - | - | - | - | - | - | - | - | - | - | - | - | - | - | - | - | - | - | - |
| | Mechanic | - | - | - | - | - | - | - | - | - | - | - | - | - | - | - | - | - | - | - | 0.2 | - | - | - | - |
| | Pilot | - | - | - | - | - | - | - | - | - | - | - | - | - | - | - | - | - | - | - | - | - | - | - | - |
| ISFP | Artist | - | - | - | - | 0.2 | 0.8 | - | - | - | - | - | 0.1 | - | - | - | - | - | - | - | - | - | - | - | - |
| | Musician | - | - | - | - | - | 0.6 | - | - | - | - | **1.0** | 0.2 | - | 0.1 | - | - | - | - | - | 0.1 | - | - | - | - |
| | Photographer | - | - | - | - | - | 0.6 | - | - | - | - | **1.0** | 0.3 | - | 0.1 | - | - | - | 0.1 | - | 0.1 | - | - | - | - |
| INFP | Writer | - | - | - | 0.2 | **1.0** | **1.0** | - | - | - | - | - | 0.3 | 0.5 | 0.9 | - | 0.1 | - | - | - | 0.1 | - | - | - | - |
| | Counselor | - | - | - | - | **1.0** | **1.0** | - | - | - | - | - | 0.1 | **1.0** | 0.9 | - | - | - | 0.1 | - | - | - | - | - | - |
| | Social Worker | - | - | - | 0.1 | **1.0** | **1.0** | - | - | - | - | - | - | 0.7 | 0.9 | - | - | - | - | - | - | - | - | - | - |
| INTP | Scientist | - | - | - | - | - | - | - | - | - | - | - | - | - | - | - | - | - | - | - | 0.1 | - | - | - | - |
| | Software Dev. | - | - | - | - | - | - | - | - | - | - | - | - | - | - | - | - | - | - | - | - | - | - | - | - |
| | Mathematician | - | - | - | - | - | - | - | - | - | - | - | - | - | - | - | - | - | - | - | - | - | - | - | 0.1 |
| ESTP | Entrepreneur | - | - | - | - | **1.0** | **1.0** | - | 0.6 | - | - | **1.0** | **1.0** | **1.0** | 0.8 | - | - | - | - | - | 0.1 | - | - | 0.1 | - |
| | Salesperson | - | - | - | - | **1.0** | **1.0** | - | 0.7 | - | - | 0.7 | 0.7 | **1.0** | 0.8 | - | - | - | - | 0.1 | - | - | - | - | 0.3 |
| | Athlete | - | - | - | - | **1.0** | **1.0** | 0.6 | 0.3 | - | - | 0.5 | 0.7 | **1.0** | **1.0** | - | - | - | - | 0.1 | 0.1 | - | - | - | - |
| ESFP | Actor | - | - | - | - | **1.0** | 0.9 | - | - | - | - | - | 0.2 | 0.5 | 0.3 | - | - | - | - | - | - | - | - | - | - |
| | Singer | - | - | - | - | **1.0** | 0.9 | - | - | - | - | - | 0.1 | - | 0.4 | - | 0.1 | - | - | - | 0.1 | - | - | - | - |
| | Event Planner | - | - | - | - | **1.0** | **1.0** | - | - | - | - | - | - | 0.9 | - | - | - | - | - | - | - | - | - | - | - |
| ENFP | Journalist | - | - | - | 0.7 | **1.0** | **1.0** | **1.0** | **1.0** | - | - | **1.0** | **1.0** | **1.0** | **1.0** | - | 0.2 | - | 0.1 | - | - | **1.0** | **1.0** | **1.0** | **1.0** |
| | Teacher | - | - | - | 0.6 | **1.0** | **1.0** | **1.0** | **1.0** | - | - | **1.0** | **1.0** | **1.0** | **1.0** | - | 0.2 | - | 0.2 | - | - | **1.0** | **1.0** | **1.0** | **1.0** |
| | Actor | - | - | - | 0.8 | **1.0** | **1.0** | **1.0** | **1.0** | - | - | **1.0** | **1.0** | **1.0** | **1.0** | - | - | - | 0.2 | - | - | **1.0** | **1.0** | **1.0** | **1.0** |
| ENTP | Inventor | - | - | - | - | **1.0** | **1.0** | **1.0** | **1.0** | - | - | **1.0** | 0.8 | **1.0** | **1.0** | - | 0.1 | - | - | - | 0.1 | - | - | - | - |
| | Entrepreneur | - | - | - | 0.3 | **1.0** | **1.0** | **1.0** | **1.0** | - | - | **1.0** | 0.8 | **1.0** | **1.0** | - | - | - | - | - | - | - | - | - | - |
| | Lawyer | - | - | **1.0** | 0.3 | **1.0** | 0.8 | **1.0** | **1.0** | - | - | **1.0** | 0.6 | **1.0** | 0.8 | - | - | - | - | - | - | - | - | - | - |
| ESTJ | Manager | - | - | - | 0.7 | **1.0** | 0.9 | **1.0** | 0.8 | - | - | **1.0** | **1.0** | **1.0** | **1.0** | - | - | - | - | 0.2 | 0.2 | - | - | - | 0.3 |
| | Military Officer | - | - | **1.0** | 0.9 | **1.0** | **1.0** | - | 0.8 | - | 0.1 | **1.0** | **1.0** | **1.0** | 0.8 | - | 0.2 | - | - | 0.4 | 0.2 | 0.4 | 0.6 | - | 0.4 |
| | Police Officer | - | - | **1.0** | 0.7 | **1.0** | 0.9 | - | 0.8 | - | - | **1.0** | 0.9 | **1.0** | 0.9 | - | 0.3 | - | - | **1.0** | 0.2 | 0.7 | 0.8 | - | - |
| ESFJ | Teacher | - | - | - | - | **1.0** | **1.0** | - | 0.1 | **1.0** | 0.3 | - | 0.3 | - | 0.3 | - | 0.2 | 0.6 | 0.3 | - | 0.3 | - | - | - | - |
| | Nurse | - | - | - | - | **1.0** | **1.0** | - | 0.1 | **1.0** | 0.3 | - | 0.2 | - | 0.1 | - | 0.1 | - | 0.4 | - | 0.3 | - | - | - | - |
| | Event Planner | - | - | - | 0.1 | **1.0** | **1.0** | - | - | **1.0** | 0.2 | - | 0.2 | - | - | - | 0.1 | **1.0** | 0.2 | - | - | - | - | - | - |
| ENFJ | Teacher | 0.7 | 0.6 | **1.0** | **1.0** | **1.0** | **1.0** | **1.0** | **1.0** | - | 0.7 | **1.0** | **1.0** | **1.0** | **1.0** | **1.0** | 0.7 | **1.0** | **1.0** | - | - | **1.0** | **1.0** | **1.0** | **1.0** |
| | Counselor | 0.5 | 0.7 | **1.0** | **1.0** | **1.0** | **1.0** | **1.0** | **1.0** | - | **1.0** | **1.0** | **1.0** | **1.0** | **1.0** | **1.0** | 0.8 | 0.9 | **1.0** | - | - | **1.0** | **1.0** | **1.0** | **1.0** |
| | HR Manager | 0.3 | 0.4 | **1.0** | **1.0** | **1.0** | 0.8 | **1.0** | **1.0** | - | 0.5 | **1.0** | **1.0** | **1.0** | **1.0** | **1.0** | 0.8 | **1.0** | 0.9 | - | - | **1.0** | **1.0** | **1.0** | **1.0** |
| ENTJ | CEO | - | - | **1.0** | **1.0** | - | 0.2 | **1.0** | 0.9 | - | - | **1.0** | 0.8 | **1.0** | 0.9 | - | 0.1 | - | - | **1.0** | - | - | - | **1.0** | 0.4 |
| | Military Officer | - | - | **1.0** | 0.8 | **1.0** | 0.9 | **1.0** | 0.9 | - | 0.1 | **1.0** | 0.7 | **1.0** | 0.7 | - | - | - | - | **1.0** | 0.1 | - | - | **1.0** | 0.3 |
| | Politician | - | - | **1.0** | 0.4 | - | 0.4 | **1.0** | **1.0** | - | - | **1.0** | 0.7 | **1.0** | 0.9 | - | - | - | - | **1.0** | - | - | - | 0.4 | 0.6 |

Table 7: (**RQ3**) Average accuracy results on the Role&Personality-Conditioned MBTI test, for all models, temperature settings, and choices of personality type and role. (To avoid cluttering, a hyphen is used in place of 0.0)

| ↑ Factor | Model / Temp. ($\tau$) / Role | Mixtral 0.01 | Mixtral 0.70 | Llama2-13 0.01 | Llama2-13 0.70 | SOLAR 0.01 | SOLAR 0.70 | Llama3-8 0.01 | Llama3-8 0.70 | Mistral 0.01 | Mistral 0.70 | NeuralChat 0.01 | NeuralChat 0.70 | Dolphin 0.01 | Dolphin 0.70 | Vicuna 0.01 | Vicuna 0.70 | Llama2-7 0.01 | Llama2-7 0.70 | Falcon 0.01 | Falcon 0.70 | Gemma 0.01 | Gemma 0.70 | Phi3 0.01 | Phi3 0.70 |
|---|---|---|---|---|---|---|---|---|---|---|---|---|---|---|---|---|---|---|---|---|---|---|---|---|---|
| Extrav. | Salesperson | 0.0 | -1.1 | 3.8 | 14.1 | 41.5 | 41.3 | -2.9 | 11.3 | 0.0 | 2.1 | 26.6 | 29.4 | 37.5 | 27.2 | 23.6 | -1.3 | 16.9 | 1.5 | 0.0 | 1.6 | 31.5 | 26.9 | 16.8 | 14.7 |
| | Actor | 0.0 | 0.1 | 3.8 | 13.5 | 44.1 | 35.8 | 6.3 | 15.7 | 0.0 | -0.9 | 27.7 | 25.6 | 37.5 | 32.3 | 11.0 | 4.3 | 2.3 | 3.5 | -7.1 | -5.1 | 29.5 | 24.9 | 16.8 | 17.8 |
| | Politician | 0.0 | -0.7 | 3.8 | 15.8 | 47.4 | 43.5 | -4.0 | 18.2 | 0.0 | -3.9 | 26.0 | 27.7 | 39.0 | 35.6 | 18.9 | -6.3 | 2.3 | 3.8 | 0.0 | 1.3 | 29.2 | 24.6 | 15.0 | 16.9 |
| Agreea. | Nurse | 0.3 | 0.5 | -8.6 | -5.2 | 40.7 | 35.8 | 7.1 | 12.0 | 0.0 | -0.9 | 22.2 | 27.2 | 33.3 | 14.2 | -2.3 | 2.3 | -6.3 | -17.1 | 0.0 | -3.9 | 12.4 | 12.7 | 7.1 | 6.5 |
| | Teacher | 0.8 | 1.0 | -6.8 | -2.9 | 40.7 | 32.5 | 7.1 | 14.0 | 0.0 | 0.7 | 22.2 | 27.5 | 30.3 | 19.1 | 0.0 | 3.4 | 4.3 | -15.6 | -2.1 | 0.4 | 12.4 | 14.2 | 5.7 | 6.2 |
| | Social Worker | -0.8 | 0.3 | -2.3 | -1.0 | 40.7 | 35.5 | 7.1 | 10.2 | 0.0 | 0.7 | 22.2 | 24.3 | 27.9 | 18.3 | -4.7 | 5.7 | -6.8 | -19.6 | -1.2 | -2.9 | 14.5 | 14.0 | 7.1 | 6.7 |
| Consc. | Accountant | 0.0 | -2.6 | 25.7 | 16.5 | 30.8 | 31.0 | 0.0 | 8.7 | 0.0 | -7.8 | 25.0 | 32.6 | 43.2 | 31.4 | 18.5 | 7.7 | 31.1 | 11.8 | 0.0 | -9.3 | 11.8 | 10.8 | 21.6 | 15.7 |
| | Software Dev. | 0.0 | 0.1 | 25.7 | 16.0 | 32.0 | 31.9 | 0.0 | 6.8 | 0.0 | -1.2 | 25.0 | 32.0 | 43.2 | 25.9 | 19.1 | 13.0 | 29.3 | 11.5 | 0.0 | 3.6 | 11.8 | 11.3 | 21.6 | 15.7 |
| | Engineer | 0.0 | -0.5 | 25.7 | 16.0 | 26.6 | 31.9 | 0.0 | 8.7 | 0.0 | -8.4 | 25.0 | 32.6 | 40.0 | 30.5 | 18.5 | 7.7 | 27.8 | 11.0 | 0.0 | -1.3 | 11.8 | 11.6 | 17.8 | 15.7 |
| Neuro. | Comedian | 0.0 | 2.3 | 44.0 | 38.8 | 40.0 | 48.9 | 125.0 | 122.5 | 0.0 | -1.0 | 61.9 | 47.5 | 29.5 | 35.9 | 66.7 | 11.3 | 110.0 | 41.4 | 7.7 | -1.2 | 98.0 | 88.1 | 131.2 | 96.3 |
| | Artist | 0.0 | 0.6 | 60.0 | 74.1 | 44.0 | 53.0 | 232.5 | 142.2 | 0.0 | 0.1 | 60.0 | 46.6 | 39.1 | 38.5 | 61.7 | 12.0 | 50.0 | 33.8 | 7.7 | 0.8 | 103.4 | 85.3 | 131.2 | 101.6 |
| | Journalist | 0.0 | -1.7 | 44.0 | 63.6 | 44.0 | 60.3 | 176.7 | 140.9 | 0.0 | -3.4 | 61.9 | 50.7 | 32.3 | 31.1 | 66.7 | 37.3 | 50.0 | 45.1 | 7.7 | -0.4 | 107.6 | 86.9 | 112.5 | 88.1 |
| Openn. | Scientist | 0.0 | -0.6 | 13.0 | 12.0 | 25.0 | 26.2 | 42.6 | 36.0 | 0.0 | 3.4 | 18.4 | 23.4 | 20.0 | 22.9 | 51.2 | 11.8 | -2.1 | 8.7 | 5.0 | -8.2 | 17.5 | 19.4 | 6.7 | 5.9 |
| | Professor | 0.0 | -0.6 | 13.0 | 12.5 | 25.0 | 23.6 | 42.6 | 32.9 | 0.0 | 4.5 | 21.1 | 22.0 | 22.5 | 18.3 | 56.2 | 22.0 | -4.2 | 3.6 | 5.0 | -0.7 | 17.5 | 17.2 | 8.0 | 7.0 |
| | Musician | 0.0 | -0.3 | 13.0 | 11.1 | 25.0 | 26.5 | 42.6 | 31.3 | 0.0 | 2.7 | 23.7 | 26.9 | 25.0 | 23.4 | 46.2 | 16.9 | -9.2 | 2.7 | 5.0 | -5.7 | 17.5 | 17.4 | 11.1 | 9.6 |

Table 8: (**RQ3**) Percentage increase results on the Role&Personality-Conditioned BFI test w.r.t. the unconditioned BFI test, for all models, temperature settings, and choices of personality factor and role.

# K MBTI Personality Descriptions and Traits (from www.myersbriggs.org)

**ISTJ Personality Traits**

**General Traits:** Quiet, serious, earn success by being thorough and dependable. Practical, matter-of-fact, realistic, and responsible. Decide logically what should be done and work toward it steadily, regardless of distractions. Take pleasure in making everything orderly and organized—their work, their home, their life. Value traditions and loyalty.

**Strengths:** Dependable and systematic, enjoy working within clear systems and processes. Tend to be traditional, task-oriented and decisive.

**Potential development areas:** Can become set in their ways and can sometimes be seen as rigid and impersonal.

**Typical characteristics:** Thorough, conscientious, realistic, detached, analytical, observant, practical, logical, factual, efficient, systematic, organized, reserved.

**Careers & career ideas:** Like to have clear goals and realistic deadlines, and to work with factual data to solve problems and monitor progress. They prefer to work in traditional organisational environments, with people who take their responsibilities seriously. Attractive occupations tend to be within management or administrative positions, with law-enforcement and accounting also holding appeal.

**Under stress:** Will tipically become stressed in the following cases: challenging my bottom-line approach, abandoning/deviating from routine, being rushed, disregarging my established rules and regulations, noise, mess, disorder, broad information, change, uncertainty, denying personal needs, dismissing my logical decisions. Stress triggers can be things that challenge their natural preference for structure and logic. In extreme circumstances they may become accusatory and pessimistic, tending to withdraw and shut down

**Relationships:** Is generally perceived by others as someone who values traditions and is also consistent and orderly. Develop strong loyalty in relationships in their lives and they work hard to fulfill commitments.

**ISFJ Personality Traits**

**General Traits:** Quiet, friendly, responsible, and conscientious. Committed and steady in meeting their obligations. Thorough, painstaking, and accurate. Loyal, considerate, notice and remember specifics about people who are important to them, concerned with how others feel. Strive to create an orderly and harmonious environment at work and at home.

**Strengths:** Patient individual who apply common sense and experience to solving problems for other people. Responsible, loyal and traditional, enjoying serving the needs of others and providing practical help.

**Potential development areas:** May be overly cautious, and might not consider the logical consequences of their decisions. They can lack assertiveness, and risk basing their decisions on what they think will please others.

**Typical characteristics:** Dependable, responsible, loyal, considerate, sensitive, thorough, organized, practical, detailed, kind, patient, realistic, understanding.

**Careers & career ideas:** Enjoy a sense of belonging at work and like to work with people who care about and support each other. Attractive jobs are those that reward loyalty and a sense of duty, including careers in healthcare, secretarial roles and social work.

**Under stress:** Will typically become stressed in the following cases: procrastination, last-minute changes, disregarding my established rules and regulations, not being appreciated for the daily help I give, workplace conflict, others inadequacy affecting my work, noise, indecision, dismissing how I feel, insufficient time to prepare, others repeating, mistakes. In extreme circumstances they tend to be accusatory and pessimistic, tending to think the worst and shut down.

**Relationships:** Generally dependable and also committed to the partner and the groups associated with. They honour commitments and like to preserve traditions, tend to be good “caretaker”.

### INFJ Personality Traits

**General Traits:** Seek meaning and connection in ideas, relationships, and material possessions. Want to understand what motivates people and are insightful about others. Conscientious and committed to their firm values. Develop a clear vision about how best to serve the common good. Organized and decisive in implementing their vision.

**Strengths:** Enjoy finding a shared vision for everyone, inspiring others and devising new ways to achieve the vision.

**Potential development areas:** May come across as individualistic, private and perhaps mysterious to others, and may do their thinking in a vacuum, resulting in an unrealistic vision that is difficult to communicate.

**Typical characteristics:** Visionary, imaginative, reflective, compassionate, idealistic, intense, insightful, caring, contemplative, reserved, empathetic, sensitive.

**Careers & career ideas:** Enjoy working for organisations with a humanitarian mission and a reputation for integrity. They like designing innovative programs or services and serving people's spiritual needs. Attractive jobs include careers in teaching, social work and artistic professions.

**Under stress:** Will typically become stressed in the following cases: not being appreciated for making a difference, short-sightedness, indecisiveness, disorder, feeling misunderstood, loud noises, forced time management, negativity from others, lack of closure, inflexible work environment, ideas met with criticism, dismissing how I feel, conflict, having the routine disturbed. In these circumstances you may feel physically stressed and intensely angry, with an obsessive focus on certain details and a tendency to overindulge.

**Relationships:** Have a gift to intuitively understand human relationships and complex meanings as well as they often understand emphatically the feelings of their partners. They are also seen as even mysterious by others as they tend to share their internal intuitions only with those they truly trust.

### INTJ Personality Traits

**General Traits:** Have original minds and great drive for implementing their ideas and achieving their goals. Quickly see patterns in external events and develop long-range explanatory perspectives. When committed, organize a job and carry it through. Skeptical and independent, have high standards of competence and performance—for themselves and others.

**Strengths:** Often able to define a compelling, long-range vision, and can devise innovative solutions to complex problems

**Potential development areas:** May come across as cold and distant when focusing on the task in hand. They can neglect to recognise and appreciate the contributions of others.

**Typical characteristics:** Innovative, independent, logical, objective, insightful, demanding, competent, productive, theoretical, strategic, reflective conceptual.

**Careers & career ideas:** Tend to enjoy being challenged intellectually and working in an environment that is hard-driving and achievement-oriented. They relish the opportunity to work with people who are experts in their field. Appealing careers include those in scientific or technical industries such as engineering, computing or law.

**Under stress:** Will typically become stressed in the following cases: disorganized work environment, short-sightedness, micromanaging, not having a goal in mind, lack of initiative, limited time to change plans, procrastination, talking about feelings, feeling the competence challenged, indecision, dismissing logical decision, mindless rule followers. In these circumstances you may feel physically stressed and intensely angry, with an obsessive focus on certain details and a tendency to overindulge.

**Relationships:** Might find it difficult to engage in social conversations and they tend to be seen as private and reserved. They also might fail to give as much praise or intimate rapport as those around them would desire.

## ISTP Personality Traits

**General Traits:** Tolerant and flexible, quiet observers until a problem appears, then act quickly to find workable solutions. Analyze what makes things work and readily get through large amounts of data to isolate the core of practical problems. Interested in cause and effect, organize facts using logical principles, value efficiency.

**Strengths:** Tend to enjoy learning and perfecting a craft through their patient application of skills. They can remain calm while managing a crisis, quickly deciding what needs to be done to solve the problem.

**Potential development areas:** Risk focusing so much on what needs to be done immediately that they fail to see the big picture. They don't always follow through on projects that require them to work closely with others.

**Typical characteristics:** Realistic, Trouble-shooter, factual, expedient, detached, objective, adaptable, logical, independent, analytical emergent, practical

**Careers & career ideas:** Like analysing problems and responding to crises. They enjoy working autonomously, and tend to prefer hands-on or analytical work. Jobs might include surgery, agriculture or engineering.

**Under stress:** Will typically become stressed in the following cases: inefficiency, lack of Independence, inability to logically assess situations, challenging my bottom-line approach, out-of-control emotion, noise, forcing a decision, disregarding my practical realities, forced into extraverted activities, dismissing my analysis of a problem, small talk, strict guidelines.In extreme circumstances they will tend to feel alienated and upset, and prone to whingeing and hypersensitivity.

**Relationships:** Egalitarian and generally tolerant of wide ranges of behaviour, but they can surprise others around them by voicing their firm judgements when logical principles are attacked. Can be a challenge to read as they tend to be quiet and reserved.

## ISFP Personality Traits

**General Traits:** Quiet, friendly, sensitive, and kind. Enjoy the present moment, what's going on around them. Like to have their own space and to work within their own time frame. Loyal and committed to their values and to people who are important to them. Dislike disagreements and conflicts; don't force their opinions or values on others.

**Strengths:** Enjoy providing practical help or service to others, as well as bringing people together and facilitating and encouraging their cooperation.

**Potential development areas:** Because they tend to be less assertive than some types, may have less influence in the workplace, and their concern for others could prevent them from making tough decisions. They sometimes put off making decisions, in the hope that a better opportunity will come along.

**Typical characteristics:** Practical, caring, accommodating, kind, considerate, spontaneous, cooperative, observant, tolerant, modest, adaptable, gentle, loyal.

**Careers & career ideas:** Enjoy working at something that is personally meaningful. They like to work in an environment with supportive co-workers who care about one another, and may shy away from outright competition. Are likely to be attracted to jobs in healthcare, service industries and clerical professions.

**Under stress:** Will typically become stressed in the following cases: environments neglecting personal values, disruptiveness, disregarding my practical realities, conflict situations, too much happening, dismissing what I feel, lack of understanding, time pressure, procedures limiting my freedom.In these circumstances they tend to become cynical, depressed, aggressive and prone to acute self-doubt.

**Relationships:** Prizes the freedom to follow their own path and they enjoy having their own space and setting their own timetables, which they will also give to their partners. Can be difficult to know well but they can care deeply about others which they show through actions rather than words.

## INFP Personality Traits

**General Traits:** Idealistic, loyal to their values and to people who are important to them. Want to live a life that is congruent with their values. Curious, quick to see possibilities, can be catalysts for implementing ideas. Seek to understand people and to help them fulfill their potential. Adaptable, flexible, and accepting unless a value is threatened.

**Strengths:** Enjoy devising creative solutions to problems, making moral commitments to what they believe in. They enjoy helping others with their growth and inner development to reach their full potential.

**Potential development areas:** May struggle to speak up in meetings, leading others to believe they don't care or have nothing to contribute. They risk failing to convince others of the merit of their ideas.

**Typical characteristics:** Flexible, insightful, developmental, reflective, idealistic, spontaneous, complex, empathetic, compassionate, caring.

**Careers & career ideas:** Enjoy helping others develop and learn, and express their creativity through writing or visual arts. They like doing work that has meaning, and enjoy working with people who share their values. Are likely to be attracted to professions in counselling and human development, as well as within the arts and writing.

**Under stress:** Will typically become stressed in the following cases: mundane work, time management required of me, metrics, critical, negativity from others, open disrespect, people/work impeding on individuality, being rushed, shutting down my ideas, unclear expectation, values undermined or challenged, critical, routine, decisions, disharmony, being met with criticism, crows, too many people

**Relationships:** Tends to be selective and reserved about sharing their deepest feelings and values and can be sometimes difficult to understand.

## INTP Personality Traits

**General Traits:** Seek to develop logical explanations for everything that interests them. Theoretical and abstract, interested more in ideas than in social interaction. Quiet, contained, flexible, and adaptable. Have unusual ability to focus in depth to solve problems in their area of interest. Skeptical, sometimes critical, always analytical.

**Strengths:** Think strategically and are able to build conceptual models to understand complex problems. They tend to adopt a detached and concise way of analysing the world, and often uncover new or innovative approaches.

**Potential development areas:** May struggle to work in teams, especially with others who they perceive to be illogical or insufficiently task-focused. They may have no clear sense of direction and may overlook important facts or practical details.

**Typical characteristics:** Theoretical, detached, sceptical, conceptual, analytical, innovative, independent, challenging, logical, strategic, insightful, contaminated.

**Careers & career ideas:** Tend to appreciate occupations in technical and scientific fields, and gain expert knowledge. They work best in an environment that offers time and space to concentrate without interruption, and which doesn't pressure people to work in teams or attend lots of meetings. Jobs that might appeal include architect, researcher and social scientist.

**Under stress:** Will typically become stressed in the following cases: dismissing my analysis of a problem, socializing challenging my competence, noise and interruptions, small talk, following strict guidelines, talking with people who don't listen and having to repeat myself, too many extraverted activities, being in the spotlight, not finding the logic in situations, others not understanding my ideas.In extreme circumstances you will tend to feel alienated and upset, and prone to whingeing and hypersensitivity.

**Relationships:** Can be tolerant of a wide range of behaviour of those around them, however they can fail to consider the impact on others of the way or style they express their ideas.

## ESTP Personality Traits

**General Traits:** Flexible and tolerant, take a pragmatic approach focused on immediate results. Bored by theories and conceptual explanations; want to act energetically to solve the problem. Focus on the here and now, spontaneous, enjoy each moment they can be active with others. Enjoy material comforts and style. Learn best through doing.

**Strengths:** Motivate others by bringing energy into situations. They apply common sense and experience to problems, quickly analysing what is wrong and then fixing it, often in an inventive or resourceful way.

**Potential development areas:** May have difficulty in managing their time, and may lose interest in long, complex projects. Being so focused on immediate problems may lead to them ignoring long-term systematic problems, and they may also be uncomfortable discussing or focusing on relationships.

**Typical characteristics:** Active, logical, trouble-shooter, observant, resourceful, practical, adaptable, spontaneous, realistic, analytical, outgoing, enthusiastic.

**Careers & career ideas:** Enjoy taking risks, managing crises and putting out fires. They work best surrounded by active, task-oriented people in an environment that is immediate and focused on the project. Are likely to be attracted to jobs in the protective services, agriculture, manufacturing and marketing.

**Under stress:** Will typically become stressed in the following cases: challenging my bottom-line approach, inefficiencies, disregarding my practical realities, commitments, goals not resulting from efforts, required planning, routine, quick decisions, dismissing my analysis of a problem.In these circumstances you will tend to be withdrawn, distracted and paranoid, with feelings of chronic anxiety.

**Relationships:** Truly loves life and they immerse themselves in it, their partners see them as adventurous risk-takers as well as pragmatic troubleshooters, however they might be impatient with exploration of relationships.

## ESFP Personality Traits

**General Traits:** Outgoing, friendly, and accepting. Exuberant lovers of life, people, and material comforts. Enjoy working with others to make things happen. Bring common sense and a realistic approach to their work and make work fun. Flexible and spontaneous, adapt readily to new people and environments. Learn best by trying a new skill with other people.

**Strengths:** Tend to be adaptable, friendly, and talkative. They enjoy life and being around people. This personality type enjoys working with others and experiencing new situations.

**Potential development areas:** Sometimes have trouble meeting deadlines, and do not always finish what they start. They can get easily distracted.

**Typical characteristics:** Adaptable, energetic, cooperative, playful, gregarious, resourceful, enthusiastic, observant, friendly, realistic, spontaneous, tolerant.

**Careers & career ideas:** Like to make work fun, and to create a spirit of cooperation. At work learn best by trying out a new skill alongside other people. Are often attracted to careers where their outgoing nature and attention to others can be applied, including fields such as healthcare and teaching.

**Under stress:** Will typically become stressed in the following cases: not being appreciated for the daily help I give, forcing a decision, dismissing what I feel, restrained by routine, analysis paralysis, too much abstract information, uncertain of my purpose, unable to change commitments, detailed plans. In these circumstances you will tend to be withdrawn, distracted and paranoid, with feelings of chronic anxiety.

**Relationships:** Is a big life-lover who enjoy food, clothes, animals and also the companion of people.

## ENFP Personality Traits

**General Traits:** Warmly enthusiastic and imaginative. See life as full of possibilities. Make connections between events and information very quickly, and confidently proceed based on the patterns they see. Want a lot of affirmation from others, and readily give appreciation and support. Spontaneous and flexible, often rely on their ability to improvise and their verbal fluency.

**Strengths:** Moving quickly from one project to another, are willing to consider almost any possibility and often develop multiple solutions to a problem. Their energy is stimulated by new people and experiences.

**Potential development areas:** May not follow through on decisions or projects, and risk burning out from over-committing or following every possibility.

**Typical characteristics:** Imaginative, energetic, innovative, expressive, cooperative, friendly, persuasive, emergent,spontaneous, supportive, flexible, enthusiastic.

**Careers & career ideas:** The ideal working environment is one that encourages and rewards creativity, fosters teamwork and offers opportunities to work with a variety of people, particularly in order to support and enlighten them. Tend to be attracted to jobs in coaching and development, teaching and religious callings, as well as the creative arts.

**Under stress:** Will typically become stressed in the following cases: organization at the expense of creativity, too many details, obligation, distrust, thoughtlessness, lack of enthusiasm, rudeness, forced to make decisions before ready, micromanaging, too many projects at once, long term plan, mundane task, rules over relationships, commitment.In these circumstances you will tend to be over-worried, withdrawn, tunnel-visioned and prone to extreme emotions.

**Relationships:** Keenly perceptive about people and they also experience feelings of wide ranges as well as intense emotions.

## ENTP Personality Traits

**General Traits:** Quick, ingenious, stimulating, alert, and outspoken. Resourceful in solving new and challenging problems. Adept at generating conceptual possibilities and then analyzing them strategically. Good at reading other people. Bored by routine, will seldom do the same thing the same way, apt to turn to one new interest after another.

**Strengths:** Solve problems creatively and are often innovative in their way of thinking, seeing connections and patterns within a system. They enjoy developing strategy and often spot and capitalise on new opportunities that present themselves.

**Potential development areas:** Sometimes avoid making decisions and may become excited about ideas that are not feasible because of constraints on time or resources. They may be overly challenging to others and their ideas.

**Typical characteristics:** Enthusiastic, imaginative, flexible, analytical, challenging, conceptual, enterprising, resourceful, logical, outspoken, emergent, theoretical.

**Careers & career ideas:** Prefer to work in a fast-growing, high-energy atmosphere which is characterised by autonomy and the freedom to think differently. They enjoy devising technical solutions to problems and selling new ideas and opportunities to others. Careers within a wide variety of fields will appeal, include those in the creative professions, business management, finance and engineering.

**Under stress:** Will typically become stressed in the following cases: being told to do something unstimulating, stubbornness, focusing on personal problems, dismissing my analysis of a problem, shutting down my ideas, isolation, too many details, mundane tasks, my competence is not respected, inefficiency, deadlines.In these circumstances you will tend to be over-worried, withdrawn, tunnel-visioned and prone to extreme emotions.

**Relationships:** Will enjoy a good debate, their conversational style is customarily challenging as well as stimulating. Their partner will see them as energetic and lively but also independent.

## ESTJ Personality Traits

**General Traits:** Practical, realistic, matter-of-fact. Decisive, quickly move to implement decisions. Organize projects and people to get things done, focus on getting results in the most efficient way possible. Take care of routine details. Have a clear set of logical standards, systematically follow them and want others to also. Forceful in implementing their plans.

**Strengths:** Drive themselves to reach their goal, organising people and resources in order to achieve it. They have an extensive network of contacts and are willing to make tough decisions when necessary. They tend to value competence highly.

**Potential development areas:** People tend to be so focused on the objective pursuit of their goal that they ignore the ideas or feelings of others. Situations where an intimate rapport is needed are likely to be less comfortable for them. They may not collect enough information before jumping into action, and risk missing new opportunities that are not already part of their plan.

**Typical characteristics:** Assertive, decisive, realistic, logical objective, practical, structured, pragmatic, straightforward, direct, organized, responsible, efficient.

**Careers & career ideas:** Enjoy setting clear goals and deadlines, and analysing problems logically. They work best in a stable environment with clearly defined roles and responsibilities. Careers in law-enforcement, manufacturing and applied technology.

**Under stress:** Will typically become stressed in the following cases: disregarding my established rules and regulations, uncertainty, dismissing my logical decisions, working with people who are not organized, inefficiency, indecision, lack of control, constant changes, unable to complete commitments, challenging my bottom line approach.In these circumstances you will tend to become hypersensitive, emotional, domineering and inflexible.

**Relationships:** Generally enjoys interacting with others and they take relationship roles seriously and will be keen to fulfil them responsibly. Their partners will see them as conscientious as well as dependable.

## ESFJ Personality Traits

**General Traits:** Warmhearted, conscientious, and cooperative. Want harmony in their environment, work with determination to establish it. Like to work with others to complete tasks accurately and on time. Loyal, follow through even in small matters. Notice what others need in their day-to-day lives and try to provide it. Want to be appreciated for who they are and for what they contribute.

**Strengths:** Tend to be sociable and outgoing, understanding what others need and expressing appreciation for their contributions. They collect the necessary facts to help them make a decision and enjoy setting up effective procedures.

**Potential development areas:** May be overly influenced by what they think others want when making decisions, and may find it difficult to adjust plans in response to unexpected opportunities. They risk being too accepting or deferential to those in charge.

**Typical characteristics:** Organized, supportive, outgoing, practical, cooperative, realistic, sympathetic, appreciative, warm, friendly, accepting, decisive, loyal.

**Careers & career ideas:** Work best in an environment that fosters a family-like atmosphere with friendly, caring people. They enjoy interacting closely with customers and colleagues, and communicating the value of a product, service or project. Are likely to be attracted to jobs in childcare, nursing, teaching or religious institutions.

**Under stress:** Will typically become stressed in the following cases: disregarding my established rules and regulations, isolations, lack of emotional support, unintentionally treating others badly, disputing harmony, challenges to established procedures, dismissing how I feel, uncertainty, not being appreciated for the daily help I give, regulations.In these circumstances you will tend to be pessimistic and rigid, and prone to self-doubt and insensitivity.

**Relationships:** Emotionally highly attuned to others using empathy, understanding their partner's emotional needs and concerns. Their partners view them as responsive and persuasive.

## ENFJ Personality Traits

**General Traits:** Warm, empathetic, responsive, and responsible. Highly attuned to the emotions, needs, and motivations of others. Find potential in everyone, want to help others fulfill their potential. May act as catalysts for individual and group growth. Loyal, responsive to praise and criticism. Sociable, facilitate others in a group, and provide inspiring leadership.

**Strengths:** Able to get the most out of teams by working closely with them, and make decisions that respect and take into account the values of others. They tend to be adept at building consensus and inspiring others as leaders.

**Potential development areas:** Often talk a lot, and may become discouraged if they do not receive a lot of feedback from others. They expect everyone to give as much to the task as they do, and can find conflict and lack of consensus difficult to deal with. They may overlook logical, factual realities when making decisions.

**Typical characteristics:** Empathetic, diplomatic, imaginative, persuasive, organized, responsible, collaborative, enthusiastic, warm, friendly, expressive, supportive.

**Careers & career ideas:** Enjoy helping others develop new skills, structure their time and meet deadlines. They work best in an environment that promotes collaboration and harmony, especially in working towards shared goals. Are likely to be attracted to careers in counselling, teaching, healthcare or religion.

**Under stress:** Will typically become stressed in the following cases: working in uncooperative environments, seclusion, harmony is disrupted, indecision, no time for brainstorming, dismissing how I feel, procrastinators, excessive criticism, not being appreciated, short-sightedness, unexpected changes.In these circumstances you will tend to be pessimistic and rigid, and prone to self-doubt and insensitivity.

**Relationships:** Tends to focus on encouraging the growth of others around them and they quickly understand their emotional needs. Their partners will see them as gracious, expressive and congenial.

## ENTJ Personality Traits

**General Traits:** Frank, decisive, assume leadership readily. Quickly see illogical and inefficient procedures and policies, develop and implement comprehensive systems to solve organizational problems. Enjoy long-term planning and goal setting. Usually well informed, well read, enjoy expanding their knowledge and passing it on to others. Forceful in presenting their ideas.

**Strengths:** See the big picture and think strategically about the future. They are able to efficiently organise people and resources in order to accomplish long-term goals, and tend to be comfortable with taking strong leadership over others.

**Potential development areas:** May overlook the contributions of others and may neglect to consider the needs of the people who implement their plans. Because they drive themselves strongly, they risk driving others as hard and may intimidate people with their take

**Typical characteristics:** Strategic, questioning, theoretical, confident, competent, assertive, innovative, structured, challenging, direct, logical, objective, decisive.

**Careers & career ideas:** Aren't afraid to make tough decisions to move forward, and enjoy solving system-level problems. They work best in a fast-growing environment which fosters competition, rewards achievement and offers them continual new challenges. Are typically attracted to positions of leadership and management, where tough-minded analysis is key.

**Under stress:** Will typically become stressed in the following cases: misinformation, inefficiency, other challenging my competence, disregarding my logical decisions, indecisiveness, lack of control, inability to make decisions, others ignoring established guidelines, loneliness, short. sightedness, disorganization

**Relationships:** Energised by stimulating interactions with people, which they really enjoy. Is seen by their partner as decisive and fair.

# L BFI Personality Descriptions and Traits (from (John et al., 2008))

**Agreeableness Personality Factor**

**Verbal labels:** Agreeableness, Altruism, Affection.

**Conceptual definition:** Contrasts a prosocial and communal orientation toward others with antagonism and includes traits such as altruism, tender-mindedness, trust, and modesty.

**Behavioral examples:** Emphasize the good qualities of other people when I talk about them; Lend things to people I know (e.g., class notes, books, milk); Console a friend who is upset.

**Conscientiousness Personality Factor**

**Verbal labels:** Conscientiousness, Constraint, Control of impulse.

**Conceptual definition:** Describes socially prescribed impulse control that facilitates task- and goal-directed behavior, such as thinking before acting, delaying gratification, following norms and rules, and planning, organizing, and prioritizing tasks.

**Behavioral examples:** Arrive early or on time for appointments; Study hard in order to get the highest grade in class; Double-check a term paper for typing and spelling errors; Let dirty dishes stack up for more than one day (Reversed item: correlated negatively with the Big Five domain).

**Extraversion Personality Factor**

**Verbal labels:** Extraversion, Energy, Enthusiasm.

**Conceptual definition:** Implies an energetic approach toward the social and material world and includes traits such as sociability, activity, assertiveness, and positive emotionality.

**Behavioral examples:** Approach strangers at a party and introduce myself; Take the lead in organizing a project; Keep quiet when I disagree with others (Reversed item: correlated negatively with the Big Five domain).

**Neuroticism Personality Factor**

**Verbal labels:** Neuroticism, Negative Emotionality, Nervousness.

**Conceptual definition:** Contrasts emotional stability and even-temperedness with negative emotionality, such as feeling anxious, nervous, sad, and tense.

**Behavioral examples:** Accept the good and the bad in my life without complaining or bragging (Reversed item: correlated negatively with the Big Five domain); Get upset when somebody is angry with me; Take it easy and relax (Reversed item: correlated negatively with the Big Five domain).

**Openness Personality Factor**

**Verbal labels:** Openness, Originality, Open-Mindedness.

**Conceptual definition:** Describes the breadth, depth, originality, and complexity, of an individual's mental and, experiential life.

**Behavioral examples:** Take the time to learn something simply for the joy of learning; Watch documentaries or educational TV; Come up with novel setups for my living space; Look for stimulating activities that break up my routine.